\documentclass{article}

\PassOptionsToPackage{numbers, compress}{natbib}

\usepackage[final]{neurips_data_2021}

\usepackage[utf8]{inputenc} 
\usepackage[T1]{fontenc}    
\usepackage{hyperref}       
\usepackage{url}            
\usepackage{booktabs}       
\usepackage{amsfonts}       
\usepackage{nicefrac}       
\usepackage{microtype}      
\usepackage{xcolor}         
\usepackage[pdftex]{graphicx}
\usepackage{dirtytalk}
\usepackage{epstopdf}

\title{Whole Brain Vessel Graphs:\\ A Dataset and Benchmark for Graph Learning and Neuroscience (VesselGraph)}
\author{
  Johannes C. Paetzold\dag\ddag, Julian McGinnis\dag, Suprosanna Shit\dag, Ivan Ezhov\dag, Paul Büschl\dag,\\
  \textbf{Chinmay Prabhakar*, Mihail I. Todorov\ddag, Anjany Sekuboyina*,  Georgios Kaissis\dag,}\\
  \textbf{Ali Ertürk\ddag, Stephan Günnemann\dag, Bjoern H. Menze*}\\
  \dag Technical University of Munich, \ddag Helmholtz Zentrum München, *University of Zürich\\
  \texttt{johannes.paetzold@tum.de}\\
}  
\begin{document}
\maketitle
\begin{abstract}
Biological neural networks define the brain function and intelligence of humans and other mammals, and form ultra-large, spatial, structured graphs. Their neuronal organization is closely interconnected with the spatial organization of the brain's microvasculature, which supplies oxygen to the neurons and builds a complementary spatial graph. This vasculature (or the vessel structure) plays an important role in neuroscience; for example, the organization of (and changes to) vessel structure can represent early signs of various pathologies, e.g. Alzheimer's disease or stroke. Recently, advances in tissue clearing have enabled whole brain imaging and segmentation of the entirety of the mouse brain's vasculature.
Building on these advances in imaging, we are presenting an extendable dataset of whole-brain vessel graphs based on specific imaging protocols. Specifically, we extract vascular graphs using a refined graph extraction scheme leveraging the volume rendering engine \textit{Voreen} and provide them in an accessible and adaptable form through the \textit{OGB} and \textit{PyTorch Geometric} dataloaders. Moreover, we benchmark numerous state-of-the-art graph learning algorithms on the biologically relevant tasks of \textit{vessel prediction} and \textit{vessel classification} using the introduced vessel graph dataset.
Our work paves a path towards advancing graph learning research into the field of neuroscience. Complementarily, the presented dataset raises challenging graph learning research questions for the machine learning community, in terms of incorporating biological priors into learning algorithms, or in scaling these algorithms to handle sparse,spatial graphs with millions of nodes and edges.\footnote{All datasets and code are available for download at \url{https://github.com/jocpae/VesselGraph}.\\Ali Ertük, Stephan Günnemann and Bjoern H. Menze share last authorship.}

\end{abstract}
\begin{figure}[ht!]
    \centering
    \includegraphics[width=1\textwidth]{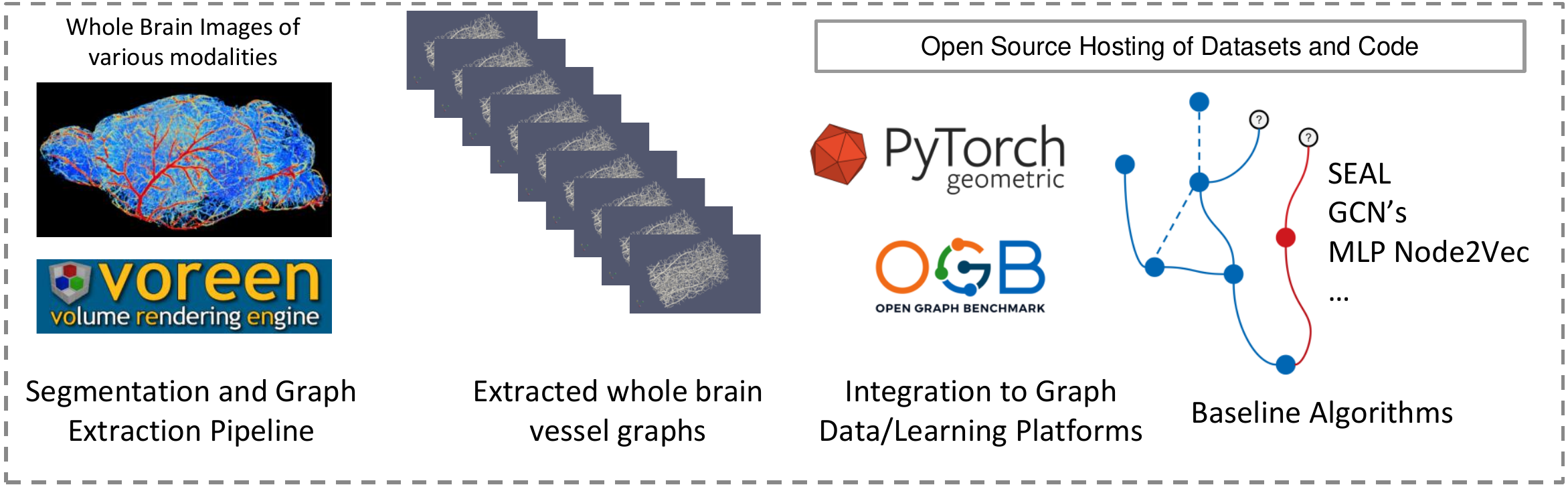}
    \caption{Graphical Abstract of \textit{VesselGraph}.}
    \label{fig:graph_abstract}
\end{figure}

\section{Introduction}
\label{Introduction}

Human intelligence and brain function are defined by the cerebral biological neuronal network, the so-called \textit{connectome}. The entirety of all single neurons forms an ultra-large, spatial, hierarchical and structured graph. Imaging and reconstructing these whole-brain graphs on a single-neuron level is one of the key problems in neuroscience. Neuronal organization is closely linked to the vascular network, as vessels supply the neurons with nutrients (e.g. oxygen). Specifically, the vessel topology determines the maximum metabolic load and determines neural growth patterns \cite{ji2021brain}. Vascular organisation, particularly in regards to vessel sizes and numbers of capillary links, varies substantially between brain regions, see Supplementary Figure \ref{fig:suppl_vessel_types} and \ref{fig:vessel_regions}. Moreover, its organization and changes to its structure are early signs for the development of specific diseases, e.g. Alzheimer's disease \cite{farkas2000pathological,bennett2018tau} or even COVID-19 encephalopathy \cite{uginet2021cerebrovascular}. As an initial step towards understanding the neuronal and vascular connectome (also known as the \textit{angiome} \cite{blinder2013cortical}), reliable imaging and segmentation methods are required. To this day, whole-brain imaging and segmentation of all neurons in the brain remains elusive. On the other hand, advances in tissue clearing and deep learning have enabled imaging and segmentation of the whole murine brain vasculature down to the microcapillary level \cite{todorov2020machine, kirst2020mapping}. 

Nonetheless, a binary segmentation of the vasculature is insufficient for a full, abstract description of the vascular connectome. To enable a comprehensive hierarchical description of the spatial vessel structure and anatomy, a graph representation of the brain with detailed features is required. This work provides the first large-scale, reproducible graph dataset thereof. 


We believe that such a graph representation can facilitate research and understanding in many fields. The correction of imperfect vascular imaging and segmentation based on such an enhanced vascular graph, could one day enable the simulation of blood-flow (hemodynamic modeling), the study of vessel anatomy, connectivity, collateralization/anastomosis and structural abnormalities. Future studies using enhanced datasets could find our approach useful to study pathologies associated with neurovascular disorders, such as stroke and dementia, given that obstacles such as plaques would be accounted for.

Evidently, the study of such spatial graphs with millions of nodes requires its own set of methods; we believe that the recent rise of advanced machine learning methods for graphs will provide suitable approaches to efficiently and accurately permit drawing deep insight from vascular graphs. This, in turn, will foster the development of methods capable of dealing with massive, but sparsely connected circular graphs, for inference on these graphs, and inference under structural and functional prior constraints that are present in such spatial physical 3D networks. 

In this work we benchmark two exemplary and biologically relevant tasks using both traditional approaches and advanced graph learning. First, in order to improve the structure and anatomical fidelity of the extracted graphs, we benchmark vessel (link) prediction. As a second task, we benchmark vessel (node) classification into the three main classes (capillaries, arterioles/venules, and arteries/veins), which represent biologically meaningful classifications by vessel size, and whose relevance for hemodynamics has been demonstrated in stroke and oxygenation modeling \cite{schmid2021severity}. 

\begin{figure}[ht]
	\centering
	\includegraphics[width=0.8\textwidth]{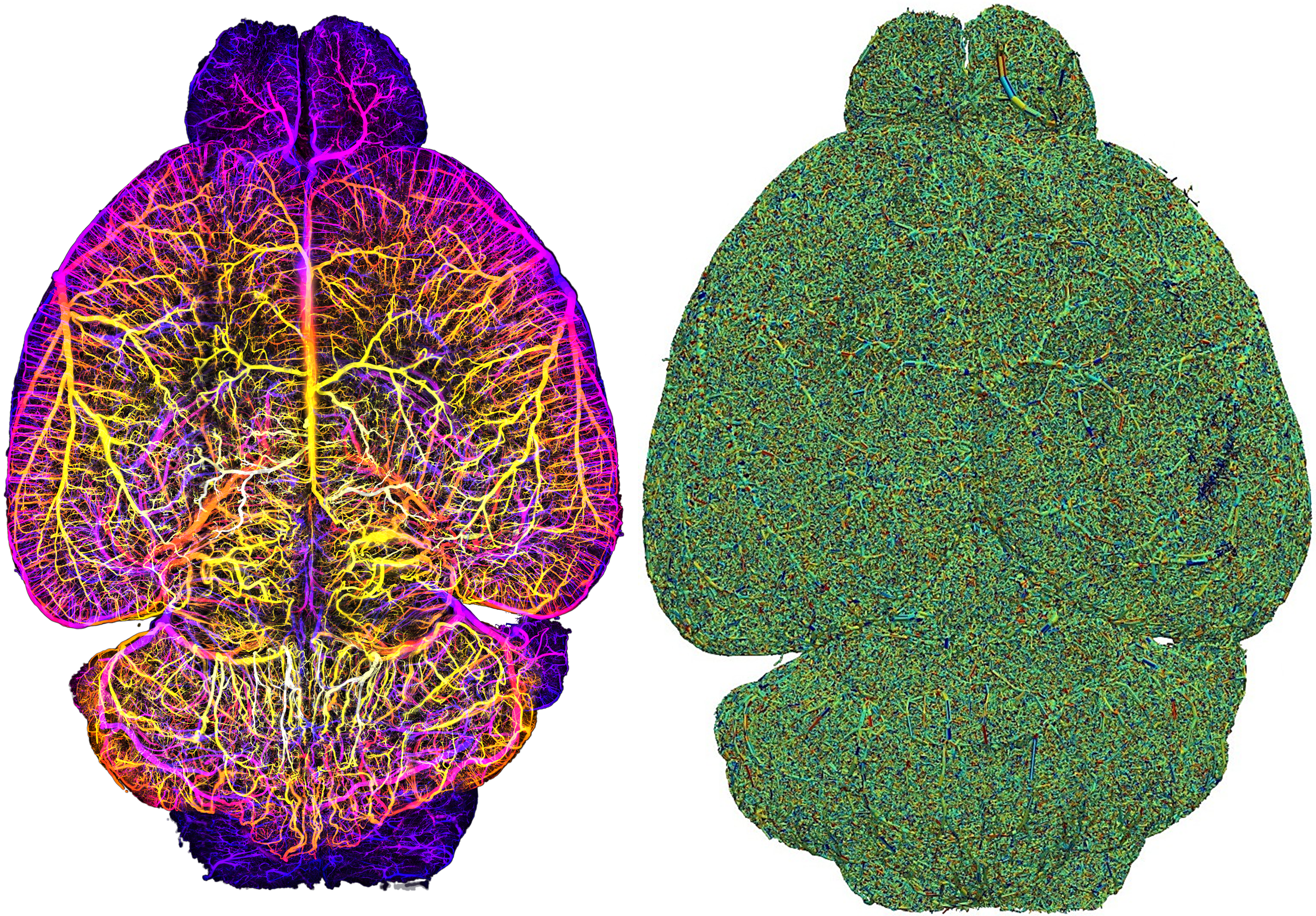}
	\caption{Left: 3D imaging of the whole mouse brain vasculature \cite{todorov2020machine} and right; the corresponding rendering of our whole brain spatial vessel graphs; the edges (vessels) are rendered with the average radius feature. 
	}
	\label{fig:3d_vessel}
\end{figure}

\subsection{Whole brain vascular imaging and segmentation}

Novel imaging methods, e.g. tissue-clearing-based methods \cite{ueda2020tissue,erturk2012three,chung2013clarity,renier2014idisco}, VesSAP \cite{todorov2020machine}, Tubemap \cite{kirst2020mapping} and the work by diGiovanna et al. \cite{di2018whole} have enabled the imaging of the full vascular structure on a whole-brain scale \cite{ji2021brain}. 

The segmentation of the resulting ultra-large and unbalanced images with thousands of pixels in each dimension (e.g. $3096 \times 4719 \times 1867$ pixels \cite{todorov2020machine}) is a challenging computer vision task which is strongly affected by technical imaging imperfections. The best-performing segmentation approaches rely on deep learning, e.g. using the U-Net architecture, and are only trained on selected, manually annotated sub-volumes of the whole brain images \cite{todorov2020machine,kirst2020mapping,ji2021brain}, leading to further imperfections in the segmentation masks. 

The process presented in our work commences with segmentations of whole-brain vessel images, for which we use publicly available data from lightsheet microscopy (VesSAP), two-photon microscopy and a synthetic blood vessel dataset. For details refer to Appendix \ref{Individual_Licenses_Data}.
In the future, we will continuously increase the dataset with whole-brain images and segmentation as they become publicly available.

\subsection{Graph learning}

Machine learning on graphs is a highly relevant research field which aims to develop efficient machine learning algorithms exploiting the unique properties of graphs, such as structure neighborhoods and the sparse representation of complex systems. Our work concerns a particularly challenging domain - spatial, structured and ultra large biological graphs. In this paper we utilize and benchmark two fundamental graph learning tasks: node classification and link prediction to study the biological properties of the vascular connectome.

A widely recognized concept for node classification is the adaption of deep learning techniques to graphs via graph convolutional networks (GCN) \cite{kipf2016semi}, a concept which was adapted and extended for many of the algorithms that we implemented, such as such as GNNs, GCNs, and GAEs \cite{kipf2016variational,zeng2019graphsaint,hamilton2017inductive,frasca2020sign,klicpera2018predict,chiang2019cluster,kong2020flag,huang2020combining,klicpera2019directional}.            
A key approach for link prediction is a so-called \textit{labeling trick} \cite{zhang2021revisiting}, which is a concept to generate sensible training data. 
The SEAL labeling trick used in our work constructs a subgraph for two candidate nodes (enclosing subgraph) and aims to learn a functional mapping on the subgraph to predict link existence \cite{zhang2018link}.

\subsection{Our contribution}

Our main contributions are: 
\begin{enumerate}
    \item We extract a set of standardized whole-brain vessel graphs based on whole, segmented murine brain images. 
    \item We publicly release said dataset in an easily accessible and adaptable format for use in graph learning benchmarking by implementing the \textit{open graph benchmark} (OGB) \cite{hu2020open} and \textit{PyTorch Geometric} data loaders \cite{fey2019fast}.
    \item In addition to our standard vessel graph, in which bifurcation points are nodes and vessels are edges, we propose an alternative representation of the vascular connectome as a line graph (where vessels become nodes), enabling the use of a multitude of advanced \textit{node classification algorithms} for vessel property prediction. 
    \item We extensively benchmark graph algorithms for the biologically important tasks of \textbf{link prediction} and \textbf{node classification}, which can serve as baselines for further research efforts in graph learning and neuroscience. 
\end{enumerate}

The rest of the paper is organized as follows: In Section \ref{Graph_Ext}, we describe our refined graph generation process and provide implementation details for the used \textit{voreen} framework and compare to other graph generation methods. We introduce the structure of our 3D brain vessel graph and provide statistics on the different extracted graphs from different brains in Section \ref{Vessel_Graph}. We describe how we generated an alternative line graph representation in Section \ref{Line_Graph}. In Section \ref{Link_Base}, we benchmark the link prediction task and in Section \ref{Node_Base}, we benchmark the node classification task on a multitude of baseline algorithms. We conclude with a focused discussion of our contribution and outline future perspectives and topics related to dataset maintenance.

\section{Graph extraction from segmentations}
\label{Graph_Ext}

Our graph extraction protocol begins with a given segmented whole-brain vascular dataset. Independent of segmentation method used (deep learning or filter-based), we tested the following state-of-the-art graph extraction algorithms: 1) the TubeMap method \cite{kirst2020mapping} which uses pruning on a 27-neighborhood skeletonization after a deep learning based tube-filling algorithm, based on a modified DeepVesselNet architecture \cite{tetteh2018deepvesselnet}; 2) the metric graph reconstruction algorithm by Aanjaneya et al. \cite{aanjaneya2011metric} which reduces linear connections of a skeleton to form a more compact and topologically correct graph and 3) the \textit{Voreen} vessel graph extraction method \cite{meyer2009voreen,drees2021scalable}. We tested the graph extraction algorithms on different imaging modalities, varying brain areas, and the synthetically generated vascular trees \cite{schneider2012tissue}.

\begin{figure}[h!]
    \centering
    \includegraphics[width=\textwidth]{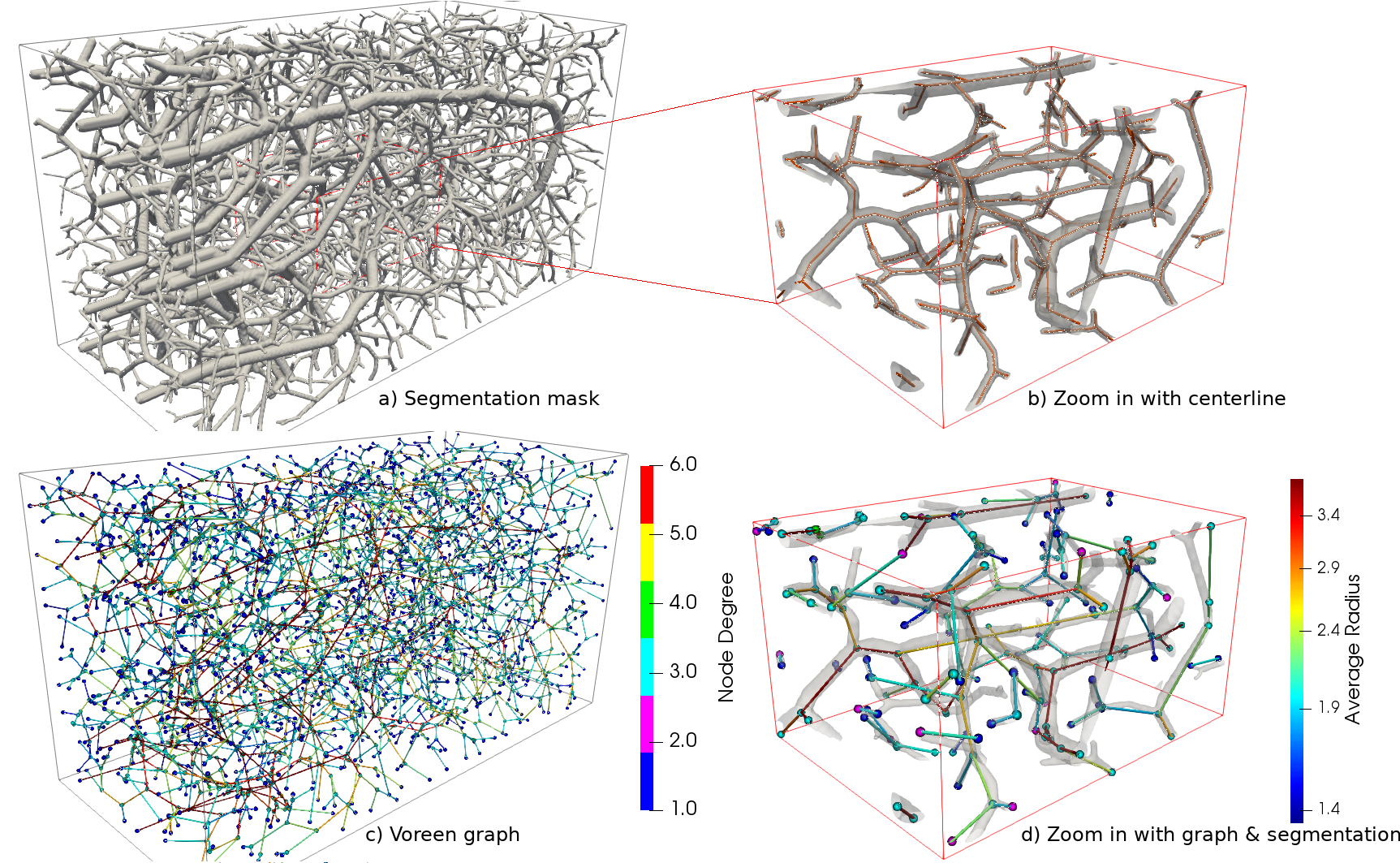}
    \caption{Extracted spatial vessel graph on a synthetic vessel volume \cite{schneider2012tissue}; the graph is extracted using the \textit{Voreen} software \cite{meyer2009voreen}; a) the original vascular segmentation rendered in rendered in grey; b) depiction of the centerlines in red for a zoomed-in section; c) the nodes with a discrete colorbar encoding their degree; d) depiction of the segmentation with the edges and a continuous colorbar encoding the radius.}
    \label{fig:my_label}
\end{figure}

After expert-level evaluation of the extracted graphs in terms of feature quality, graph robustness and pipeline parameters, and of the algorithms in terms of scalability, runtime and resource constraints, we selected \textit{Voreen} \cite{drees2021scalable} for our graph generation. For details and comparisons we refer to Supplementary section \ref{Suppl_Graph_Extraction}.

\textit{Voreen} (Volume Rendering Engine) is a general framework for multi-modal volumetric dataset visualization and analysis purposes. One key advantage of \textit{Voreen} compared to other graph generation algorithms, is that its graph extraction process is deterministic, robust and scalable. It has successfully been applied to cosmological visualization \cite{scherzinger2017interactive}, visualization of large volumetric multi-channel microscopy data \cite{brix2014visualization}, 3D visualization of the lymphatic vasculature \cite{dierkes2018three}, 3D histopathology of lymphatic malformations \cite{hagerling2017vipar} and velocity mapping of the aortic flow in mice \cite{bovenkamp2015velocity}.

Our graph extraction follows a four-stage protocol:
\begin{enumerate}
    \item Skeletonization: The binary segmentation volume is reduced to a skeleton based representation by applying a standard topological thinning algorithm by Lee et al. \cite{lee1994building}.
    \item Topology Extraction: memory efficient algorithms extract the vessel centerlines \cite{isenburg2009streaming}. \textit{Voreen} allows to store this intermediate representation in a combination with the graph. 
    \item Voxel-Branch Assignment: Computing of mapping between the so-called protograph (i.e. the initial graph) and the voxels of the binary segmentation.
    \item Feature Extraction: On basis of the protograph and the mapping, several features can be computed from the foreground segmentation.
\end{enumerate}

Multiple iterations of the four-stage protocol refine and improve the graph quality and prune small, spurious branches. The key optimization parameter for the graph structure in terms of node representation, and node statistics is the \textit{bulge size}. Expert neuroscientists determined the parameter (bulge size = 3, a parameter choice in line with previous work \cite{drees2021scalable}) by statistically comparing the resulting graphs, and visually interpreting the vascular connections in varying brain regions (compare Supplementary Figure 6). Still, known limitations of topological thinning-based methods for graph extraction exist \cite{drees2021scalable}, motivating our first baseline task, presented in Section \ref{Link_Base}. 

\section{3D vessel graph dataset}
\begin{table}[h!]
    \centering
    \footnotesize
    \begin{tabular}{ p{2.5cm}||p{2.2cm}|p{2cm}|p{2cm}}
    \hline
    \hline
    \multicolumn{4}{c}{Complete Datasets} \\
    \hline \hline
    Name & Number of Nodes & Num of Edges & Node Degree \\
    \hline
    BALBc1 \cite{todorov2020machine}& 3,538,495 & 5,345,897 & 3.02 \\ 
    BALBc2 & 3,451,306 & 5,193,775 & 3.01 \\ 
    BALBc3 & 2,850,347 & 4,097,953 & 2.88 \\ 
    C57BL/6-1  & 3,820,133 &  5,614,677 & 2.94 \\  
    C57BL/6-2  & 3,439,962 &  5,070,439 & 2.95 \\  
    C57BL/6-3  & 3,318,212 &  4,819,208 & 2.90  \\ 
    CD1-E-1 & 3,645,963 & 5,791,309 & 3.18 \\ 
    CD1-E-2 & 1,664,811 & 2,150,326 & 2.58 \\ 
    CD1-E-3 & 2,295,360 & 3,130,650 & 2.73 \\ 
    \hline
    C57BL/6-K18 \cite{ji2021brain} & 4,284,051 & 6,525,881 & 3.05 \\ 
    C57BL/6-K19 & 3,948,612 & 5,999,958  & 3.04 \\ 
    C57BL/6-K20 & 4,165,085 & 6,317,179 & 3.03 \\ 

    \hline
    Synth. Graph 1 \cite{schneider2012tissue} & 3159& 3234 & 2.05   \\
    Synth. Graph 2 & 3349 & 3421 & 2.04  \\
    Synth. Graph 3 & 3227& 3310 & 2.05 	 \\
    Synth. Graph 4 & 3178 & 3251 & 2.05  \\
    Synth. Graph 5 & 3294 & 3376 & 2.05  \\
    \hline
    
    \end{tabular}
        \caption{Total number of edges, nodes and average node degree for the different whole brain graphs.} 
    \label{tab:graph_numbers_nodes_edges}
\end{table}

Our 3D vessel dataset features 17 graphs from 2 different imaging modalities as well as 5 sets of synthetic vascular graphs. We found the smaller synthetic graphs useful for prototyping since they are smaller in size and cover all three classes of vessels (arteries, arterioles and capillaries). For all real vessel graphs, the full 3D images and binary segmentations are also publicly available. An overview of the notation used throughout the following sections alongside typical values can be found in Table \ref{tab:graph_feats}.

\subsection{Vessel graph G}
\label{Vessel_Graph}

The output of the \textit{Voreen} graph extraction pipeline represents our primary unweighted and undirected graph or \say{intuitive} vessel graph. Let this graph be denoted as $\mathcal{G}=(\mathcal{V}, \mathcal{E})$, where $\mathcal{V}$ is the set of nodes and $\mathcal{E}$ is the set of all the edges of the graph.  

\paragraph{Nodes:}
From a biological perspective, each node $n\in \mathcal{V}$ in our graph either represents end points of the vessel branches or the bifurcation of vessel branches, (see Figure \ref{fig:graphs}). Bifurcation points are the points where a larger vessel branches into two or more smaller vessels (in case of an artery) or smaller vessels merge into a large vessel (in case of a vein). The number of vessels branching from a bifurcation point defines the degree of that particular node. Bifurcation points have node degree of 3 or higher. In some cases, our graphs also have vessel endpoints, which are encoded as nodes of degree 1. Further, degree 2 nodes are generated by the graph extraction in cases when vessels exhibit a large curvature. These nodes are important to preserve the vessel curvature in its graph representation. For a statistical evaluation of the node degree please see Supplementary Figure \ref{fig:bulge}.

\begin{table}[h!]
    \centering
    \footnotesize
    \begin{tabular}{ p{1.5cm}|p{2.5cm}|p{1.9cm}|p{4.5cm} }
    \hline
    \hline
    \multicolumn{4}{c}{Feature Overview} \\
    \hline
    \hline
    Name & Feature Type & Value & Description  \\
    \hline
    $x_n$ & node feature & $[178,3096]$ * & x-coordinate  \\
    $y_n$ & node feature & $[808,4719]$ * & y-coordinate  \\
    $z_n$ & node feature & $[0,1866]$ * & z-coordinate  \\
    $a_n$ & node feature & $\{0,1\}^{71}$ & Allen mouse brain atlas region  \\
    \hline
    \hline
    $\mu^{r}_{ij}$ & edge feature & $[0.5,38.65]$ & mean of minimum radii  \\
    $\sigma^{r}_{ij}$ & edge feature & $[0.0,12.49]$ & std. of minimum radii\\
    $\mu^{\bar{r}}_{ij}$ & edge feature & $[0.79,38.65]$ & mean of average radii\\
    $\sigma^{\bar{r}}_{ij}$ & edge feature & $[0.0,11.99]$ & std. of minimum radii\\
    $\mu^{R}_{ij}$ & edge feature & $[0.91,44.12]$ & mean of maximum radii   \\
    $\sigma^{R}_{ij}$ & edge feature & $[0.0,23.64]$ & std. of minimum radii\\
    \hline
    $\mu^o_{ij}$ & edge feature & $[0.04,1.99]$ & mean of roundness   \\
    $\sigma^o_{ij}$ & edge feature & $[0.0,1.0]$ & std. of roundness\\
    
    \hline
    $l_{ij}$ & edge feature & $[2,322.81]$ & vessel length\\
    $d_{ij}$ & edge feature & $[1.77,300.36]$ & shortest distance\\
    $\rho_{ij}$ & edge feature & $[0.18,27.43]$ & curvature  \\
    $\alpha_{ij}$ & edge feature & $[0.29,1587.49]$ & mean crosssection area   \\
    \hline
    $v_{ij}$ & edge feature & $[1.0,119459]$ & Volume of vessel   \\
    $nv_{ij}$  & edge feature & $[0.0,256] \cap\mathbb{N}$ & no. of voxel in vessel   \\
    
    \hline
    $\nu^1_{ij}$ & edge feature & $[1,14]\cap\mathbb{N}$ & degree of $n_i$ of edge $e_{ij}$ \\
    $\nu^2_{ij}$ & edge feature & $[1,14]\cap\mathbb{N}$ & degree of $n_j$ of edge $e_{ij}$  \\
    
    \hline
    \hline
    \end{tabular}
    \caption{Systematic overview of the notation of the existing node and edge features in our spatial vessel graphs. All features besides the Allen brain atlas region and the node degree are spatial and extracted using Voreen, discrete ranges are given for the \textit{Balbc1} brain (* subject to imaging resolution). }
    \label{tab:graph_feats}
\end{table}

\paragraph{Node features:}
We extract two important features for the nodes of graph $\mathcal{G}$. For each node, the key features are the physical location in the coordinate space and the anatomical location in reference to the Allen brain atlas \cite{sunkin2012allen}. For the physical location feature, we denote real valued coordinates $[x_n, y_n, z_n] \in \mathbb{R}^3 \; \forall n \in \mathcal{V}$ where $[x_n, y_n, z_n]$ is the location of node $n$ in 3D space. Further, multiple prior works have shown that regional differences in vessel geometry can be observed in different brain regions \cite{ji2021brain,hahn2021large,todorov2020machine}. This motivates us to include anatomical location features for the nodes. Hence, we register the whole segmentation volume to the Allen brain atlas. Our reference Atlas uses the ontology the Allen mouse brain atlas (CCFv3 201710). We use the average template. After appropriate downsampling of the Allen brain atlas and the images, we apply a two-step-rigid and deformable registration using \textit{elastix}. Our protocol is thus identical to the Vessap paper\cite{todorov2020machine}.  Subsequently, we assign the brain region where a particular node is located in the brain atlas as anatomical node location feature, see Supplementary Figure \ref{fig:atlas_image}. Formally, the anatomical location feature $a_n=c \; \forall n \in \mathcal{V}$ if $[x_n, y_n, z_n]\in \; A_c$, where $A_c$ is the $c$\textsuperscript{th} region of the brain atlas. The atlas includes $71$ brain regions which are hierarchically clustered from $>2000$ subregions. The anatomical location feature is embedded as a one-hot encoded vector.

\paragraph{Edges:}
Each edge $e_{ij} \in \mathcal{E}$ in our graph represents vessels or vessel segments which connect two nodes $\mathcal{V}$, see Figure \ref{fig:graphs}. These edges (vessels) determine the structure of the whole brain network and represent the core aspect of our research questions. The edges exhibit the following rich set of features, which are extracted based on the shape and topology of the given segmented images.

\paragraph{Edge features:}
We extract geometric properties for each of the edges. For that, we determine the maximum diameter inscribed circle, least square reference circle, and minimum circumscribed circle on the discretized cross-section of a vessel branch and compute their radius as $\{r^k_{ij}\}, \{\bar{r}^k_{ij}\}$ and $\{R^k_{ij}\}$ where $k=1:K$ for $K$ number of cross section of the edge $e_{ij}$, respectively. From this, we compute the mean and standard deviation of the minimum, average and maximum radius for each edge $e_{ij}$ as follows. Specifically, $\mu^r_{ij}, \sigma^r_{ij}$ denotes the mean and standard deviation of minimum radius of edge $e_{ij}$. We extend the same notation for mean and standard deviation for $\{\bar{r}^k_{ij}\}$ and $\{R^k_{ij}\}$ as $\mu^{\bar{r}}_{ij}, \sigma^{\bar{r}}_{ij}, \mu^{R}_{ij}, \mu^{R}_{ij}$ respectively. We compute the roundness of each cross section as $o^k_{ij}=\frac{r^k_{ij}}{R^k_{ij}}$. We denote the mean and standard deviation of roundness as $\mu^o_{ij}$ and $\sigma^o_{ij}$, respectively. Further, we extract the vessel length $l_{ij}$, shortest distance between two nodes of an edge $d_{ij}$, curvature $\rho_{ij}=\frac{l_{ij}}{d_{ij}}$, mean cross section are $\alpha_{ij}$. Moreover, we use the degree of the nodes $n_i$ and $n_j$ for an edge $e_{ij}$ as $\nu_i$ and $\nu_j$, respectively. The complete set of edge features can be found in Table \ref{tab:graph_feats}.

\subsection{Line vessel graph L(G)}
\begin{figure}
    \centering
    \includegraphics[width=0.95\textwidth]{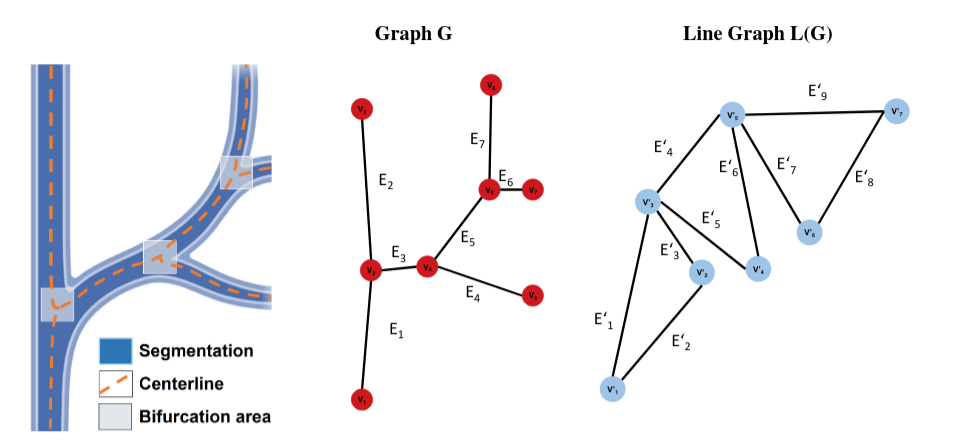}
    \caption{Depiction of an exemplary vessel tree with the the spatial vessel Graph $\mathcal{G}(\mathcal{V},\mathcal{E})$ with nodes $(\mathcal{V})$ and edges $(\mathcal{E})$; additionally, a line graph  $L(\mathcal{G})$ of the spatial vessel graph $\mathcal{G}$ ; where each node (bifurcation point) becomes an edge; two nodes of $L(\mathcal{G})$ are adjacent if and only if their edges are incident in $\mathcal{G}$.}
    \label{fig:graphs}
\end{figure}
\label{Line_Graph}
As an alternative representation of whole brain vessel graphs, we convert our vessel graphs $\mathcal{G}$ to a corresponding line-graph representation, $L(\mathcal{G})$\cite{gross2018graph}. A line graph (depicted in Figure \ref{fig:graphs}) is a graph where the edges of the base graph $\mathcal{G}$ become nodes and an edge between the new nodes is created if and only if their edges are incident in $\mathcal{E}$. Edges are the most important aspects in our graph $\mathcal{E}$ because of their one-to-one correspondence to the vessels. Therefore, we wish to apply another set of graph-learning algorithms, namely node classification algorithms, to study their biological properties based on the rich set of vessel features. Hence, we construct an alternative representation with the help of line graph $L(\mathcal{G})$. We formally define $L(\mathcal{G}) := (\mathcal{V}',\mathcal{E}')$ where $\mathcal{V}'=\mathcal{E}$ and $\mathcal{E}'=\{\{e_{ij}, e_{ik}\}$ if $\exists \; (e_{ij}, e_{ik}) \in \mathcal{E}$\}.
\paragraph{Nodes:}
Now, the nodes in the line graph $\mathcal{V}'$ represent vessels or vessel segments, see Figure \ref{fig:graphs}. 
\paragraph{Node features:}
Thus, all edge features of $\mathcal{G}$ can now be used as node features for $L(\mathcal{G})$, see Table \ref{tab:graph_feats}. One of the key advantages of constructing the line graph is that we can now leverage a large number of prior techniques presented in node classification literature such as the use of vessel features in message passing.
\paragraph{Edges:}
Edges are defined as pairwise adjacencies of two nodes (vessels) \textit{if and only if} the corresponding edges in $\mathcal{G}$ are connected to a node $\mathcal{V}$. In practice, this means that nodes in $\mathcal{G}$ which are of degree $1$ disappear in $L(\mathcal{G})$ and that each node in $\mathcal{G}$ with a degree $\geq 2$ will create multiple edges in $L(\mathcal{G})$. 
\paragraph{Edge features:}
The spatial location given as node features in $\mathcal{G}$ can now be added as an edge feature.

\section{Benchmarking link prediction}
\label{Link_Base}

The formal goal of link prediction is to train a classifier $\mathcal{F}$ which predicts links in $\mathcal{E}_{\mbox{\textit{pred}}}$ as positive and negative labels, it can be formalized as follows $\mathcal{F} :\mathcal{E}_{\mbox{\textit{pred}}} \rightarrow \{0,1\}$.

From a biological perspective this task is relevant to correct missing and imperfect vessel graph connections, because the extracted graph may be over- or under-connected, due to artifacts and shortcoming of the segmentation and network extraction.

In order to provide initial baselines for vessel (link) prediction, we implemented 10 models. The following graph learning baselines were trained without edge features: the GCN by Kipf et al. \cite{kipf2016semi}, a GNN using the GraphSAGE operator \cite{hamilton2017inductive} and the SEAL GNN, a network aiming to learn general graph structure features from the local subgraph \cite{zhang2018link}. Furthermore, we trained a multilayer perceptron (MLP) on full batches based on Node2Vec features \cite{grover2016node2vec}. Apart from these, more traditional, heuristic-based methods were implemented for the task of link prediction, which include the Katz index \cite{katz1953new}, Common Neighbour, Page Rank and Adamic Adar \cite{adamic2003friends}, a measure which computes the closeness of nodes. These traditional methods make predictions based on the graph structure itself.

\subsection{Dataset curation - SEAL}	

\paragraph{Link sampling strategy:} 
\label{Link_Sampling}

The curation of a balanced training dataset requires the introduction of two types of edges. Similar to the SEAL paper \cite{zhang2018link}, we use the notion of \textit{positive edges} and \textit{negative edges}. Generally, positive edges are random samples of existing links and negative edges are samples of non-existent links between randomly chosen nodes of the dataset (which are included in the adjacency matrix). For positive edges, we utilize random samples of the existing edges of each graph. However, since our dataset includes 3D coordinates as the node features, their spatial nature makes selecting negative samples more challenging. A trivial random selection, which has been used in other state-of-the-art methods such as SEAL, would lead to biologically implausible edges, e.g. an edge between two nodes in different brain hemispheres. These can be easily distinguished based on the coordinates and thus would not provide useful information to the model. As such, models trained with trivial random sampling struggle with the link prediction task. To address this issue, we restrict negative edge sampling to a coordinate space which spatially surrounds the source node, and choose the target node by randomly selecting nodes that are located within the following cubic space around the source node: $\delta = \overline{l_{i,j}} + 2 \sigma $, where $\overline{l_{i,j}}$ denotes the average vessel length in $\mathcal{G}$. We note that this link sampling strategy is a first baseline and could be improved upon in future work.
\paragraph{Experiment:}
For our GCN based architectures we did an extensive grid-search of hyper-parameter combinations on a subset of the whole brain graph. We subsequently trained on the whole brain graphs. This intermediate step was necessary because exploring thousands of hyper-parameter combinations on the whole brain dataset is computationally infeasible. Implementation details and details on the hyperparameter search are indicated in supplementary Table \ref{tab:link_hyper_parameters}.
	
For the main experiment we sample all edges from one whole brain graph as positive edges $\mathcal{G(V,E)}$ (BALBc-1, Vessap, see Table \ref{tab:graph_numbers_nodes_edges})and randomly assign these to the training, validation and test set (80/10/10 split). Moreover, we sample an identical number of negative edges, i.e. non-existent but theoretically probable links according to the curation criterion described above. Next, we randomly shuffle all negative edges. Thus, we mitigate any bias in the negative train, validation and test splits and ensure a region-independent distribution. Subsequently, we randomly assign the negative edges to the train, validation and test set (80/10/10 split). This provides us with a balanced datast in regards to positive and negative edges.

We choose to only use the spatial node features for our experiment: ${x_n,y_n,z_n}$. This task is very hard because the algorithm essentially has to learn the vascular graph hierarchy purely on undirected relational and spatial information. 

\begin{table}[h!]
\footnotesize
\centering
\caption{\label{tab:results_link_pred} Results for the link prediction baselines.}
\begin{tabular}{| l||c|c|}
\hline
Algorithm  & \multicolumn{2}{|c|}{ROC AUC} \\
                        & validation & test  \\
\hline
Adamic Adar & 48.49  & 48.49   \\
Common Neighbors & 48.50  & 48.49  \\
Resource Allocation &48.49 & 48.50  \\
\hline
Matrix Factorization & 50.07   & 50.08  \\
MLP  &  57.98  &  58.02 \\
\hline
GCN GCN & 50.69  &  50.72  \\
GCN GCN + embeddings&  51.32 &  51.13  \\
GCN SAGE + embeddings   & 52.81  & 52.88 \\
GCN SAGE &  59.37 &  59.23  \\ 
\hline
SEAL  &  \textbf{91.01} & \textbf{90.96} \\ 
\hline
\end{tabular}
\end{table}
Generally, traditional methods and simple GCN models performed poorly. Among the traditional methods tested, the MLP performed best. On the other hand, the SEAL implementation reached a superior performance and a strong inductive bias (ROC AUC $>90\%$). This improvement is in line with recent literature \cite{zhang2021revisiting}, which found a considerable performance improvement as a result of the employed labeling trick. This highlights that complex, dedicated graph-learning concepts need to be developed to address biologically inspired spatial graph challenges. A detailed experimental description and interpretation can be found in the Supplementary material, section \ref{Suppl_Link_Pred}.

\section{Benchmarking vessel attribute classification}
\label{Node_Base}

Our formal goal of node classification is to train a classifier $\mathcal{F}$ which predicts a class label $\mathcal{Y}$ out of a set of possible classes $\mathcal{N}_n$ of a node $\mathcal{V}$, it can be formalized as follows $\mathcal{F} :\mathcal{V} \rightarrow \mathcal{Y}\in \mathcal{N}_n$.

Biologically, this task is relevant because the vessel radius is one of the most important parameters for blood flow; any task associated with flow modelling (such as stroke diagnosis and treatment) is heavily dependent on the diameter of the affected vessel. For example in stroke, a different treatment option is chosen based on the size of the vessel in the context of its local network topology. Therefore, reliably classifying vessel segments into categories such as arteries/veins, arterioles/venules and capillaries is relevant.  

For the secondary task of vessel radius (node) classification we implemented 7 graph and non-graph learning baselines discussed in the OGB paper \cite{hu2020open}. Among them node classification using an MLP initialized on N2Vec \cite{grover2016node2vec}, a simple GCN \cite{kipf2016semi}, a GNN using the GraphSAGE operator \cite{hamilton2017inductive}, the GraphSAINT algorithm which includes a mini-batch GCN\cite{zeng2019graphsaint}, the Scalable Inception Graph Neural Networks (SIGN)\cite{frasca2020sign} and the Cluster-GCN algorithm\cite{chiang2019cluster}.
Furthermore we implemented SpecMLP-W + C\&S and SpecMLP-W + C\&S + N2Vec, which use shallow models ignoring graph structure and standard label propagation techniques from semi-supervised learning methods \cite{huang2020combining}. 

\paragraph{Experiment:}
We split our three classes according to the minimum radius feature $\mu^{r}_{ij}$ into classes of $\mu^{r}_{ij}< 15\mu m;15-40\mu m$ and $>40\mu m$. Defined by the anatomy and properties of oxygen distribution these three classes are highly imbalanced. E.g. for the \textit{Vessap} datasets the distribution is roughly $95\%$, $4\%$ and $1\%$. Similarly to the link prediction task we carried out a grid search for optimal hyper-parameters, see Supplementary Table \ref{tab:node_hyper_parameters}. We randomly split the nodes into train, validation and test sets of (80/10/10) of one whole mouse brain (BALBc-1, Vessap, see Table \ref{tab:graph_numbers_nodes_edges}). We choose to use the following node features for our experiment: $l_{ij}$ , $d_{ij}$ and $\rho_{ij}$.

\begin{table}
\centering
\footnotesize
\caption{\label{tab:results_node_prediction} Results of the implemented node classification baselines. The performance scores are the weighted F1 score, one versus rest ROC AUC, class balanced accuracy and total accuracy (ACC).}

\begin{tabular}{| l||c|c|c|c|c|c |c|c |}
\hline
  & \multicolumn{2}{|c|}{F1 Score}  & \multicolumn{2}{|c|}{ROC AUC} &                                                 \multicolumn{2}{|c|}{Balanced ACC} & \multicolumn{2}{|c|}{ACC} \\
                        & valid & test & valid & test & valid & test & valid & test\\
\hline

GCN                                         & $75.74$   & $75.75$   
                                            & $67.23$   & $66.46$
                                            & $58.38$   & $56.83$
                                            & $62.94$   & $62.92$ \\
GraphSAGE                                   & $81.98$   & $81.98$   
                                            & $\mathbf{77.35}$ & $\mathbf{77.18}$
                                            & $\mathbf{71.82}$   & $\mathbf{71.33}$
                                            & $72.02$   & $71.98$ \\
GraphSAINT                                  & $77.46$   & $77.40$   
                                            & $71.38$   & $70.71$
                                            & $63.74$   & $62.51$
                                            & $64.88$   & $64.84$ \\
SIGN                                        & $74.46$   & $74.49$   
                                            & $67.26$   & $66.04$
                                            & $57.90$   & $55.88$
                                            & $61.25$   & $61.27$ \\
Cluster-GCN                                 & $\mathbf{86.10}$   & $\mathbf{86.06}$  
                                            & $\mathbf{77.91}$   & $\mathbf{77.43}$
                                            & $\mathbf{72.23}$   & $\mathbf{71.87}$
                                            & $\mathbf{77.47}$   & $\mathbf{77.41}$ \\
\hline
MLP                                         & $76.11$   & $76.11$   
                                            & $58.08$   & $57.79$
                                            & $42.36$   & $41.72$
                                            & $63.65$   & $63.61$ \\
SpecMLP-W + C\&S                        & $84.48$   & $84.55$   
                                            & $58.12$   & $58.54$
                                            & $42.20$   & $42.93$
                                            & $75.84$   & $75.91$ \\
SpecMLP-W +  + N2Vec             & $80.53$   & $80.63$   
                                            & $66.69$   & $66.20$
                                            & $59.04$   & $57.90$
                                            & $69.99$   & $70.10$ \\
\hline
\end{tabular}
\end{table}

For node classification, we find acceptable to high performance in our baselines 
by all the methods we tested. More complex graph models such as GraphSAGE and Cluster-GCN outperform simple GCNs on average over all metrics. According to the metrics which account for class imbalance i.e. ROC AUC and balanced ACC, graph neural networks outperform non-graph learning methods, for a detailed interpretation see Supplementary \ref{Suppl_Node_Class}.

\section{Discussion}
\label{Discussion}

In this work, we introduce and make publicly available a large dataset of vessel graphs representing the most comprehensive and highest resolution representation of the whole vascular connectome to-date. We provide this set of graphs as a new \say{baseline dataset} for machine learning on graphs and make it re-usable and easily accessible by leveraging widely employed open standards, such as the \textit{OGB} and \textit{PyTorch Geometric} dataloaders. 

To provide an example for the utilization of our dataset and to promote graph machine learning research in neuroscience, we provide two benchmarks: First, we benchmark vessel (link) prediction to improve the vascular connectome; second, we implement vessel (node) classification into three main anatomical categories on the line graph. We thus show that graph learning-based methods outperform traditional methods for vessel (node) classification. Moreover, we demonstrate that link prediction based solely on the spatial organization is a difficult task for most algorithms. However, we provide evidence that the combination of an appropriately chosen, complex GNN model (SEAL) with a labeling trick can achieve high accuracy on this task, paving the way for dedicated machine learning research on spatial (biological) graphs as a key to unlocking biological insight. 

\paragraph{Dataset bias:} While the dataset and the evaluation we provide are thorough, we note the following bias in our work: Our vascular graphs are constrained by the technical bias and limitations inherent to experimental imaging, such as artifacts in the clearing protocol and physical limitations concerning the resolution and isotropy of the microscopy. All specimen imaged in this study are males. Moreover, even state-of-the-art deep learning methods for segmentation presented in literature are only trained on incomplete sets of labeled data, leading to a model bias in segmentation. Further problems can occur from the known limitations of topological thinning-based methods for graph extraction \cite{drees2021scalable}. 

\paragraph{Limitations:} 

The sum of these effects and bias can impair the usefulness of our dataset for certain, highly specialised tasks, such as flow simulations using the Navier-Stokes equations, which are strongly dependent on accurate radius measurements. 

Moreover, benchmarking all available features, data and concepts was beyond the scope of our work. For instance, an extension to heterogeneous graph representations \cite{schlichtkrull2018modelingrgcn, vashishth2019composition}, the utilization of more features, the inclusion of more than one graph or of weighted graphs, where e.g. all edges (vessels) are weighted depending on an embedding of their radius, may facilitate an improved interpretation. 
In summary, we are convinced that both the machine learning concepts and the biological insight arising from our work can be translated to other tasks, such as graph extraction and refinement on different vascular or neuronal imaging techniques, artery and vein classification, and even vessel classification in inherently different medical imaging protocols such as angiography for stroke diagnosis. We are thus hopeful that our provision of high-quality data and strong baselines will stimulate future research in this area.

\begin{ack}
We thank Dominik Drees for his advice and help with setting up the \textit{Voreen} pipeline, and Muhan Zhang for his advice on enhancing SEAL. Moreover, we would like to thank Mattias Fey, Weihua Hu and the OGB Team for providing PyTorch geometric and baseline implementations.

\end{ack}

\small{
\bibliographystyle{unsrt}
\bibliography{main}

\begin{thebibliography}{10}

\bibitem{ji2021brain}
Xiang Ji, Tiago Ferreira, Beth Friedman, Rui Liu, Hannah Liechty, Erhan Bas,
  Jayaram Chandrashekar, and David Kleinfeld.
\newblock Brain microvasculature has a common topology with local differences
  in geometry that match metabolic load.
\newblock {\em Neuron}, 109(7):1168--1187, 2021.

\bibitem{farkas2000pathological}
Eszter Farkas, Gineke~I De~Jong, Rob~AI de~Vos, ENH~Jansen Steur, and Paul~GM
  Luiten.
\newblock Pathological features of cerebral cortical capillaries are doubled in
  alzheimer’s disease and parkinson’s disease.
\newblock {\em Acta neuropathologica}, 100(4):395--402, 2000.

\bibitem{bennett2018tau}
Rachel~E Bennett, Ashley~B Robbins, Miwei Hu, Xinrui Cao, Rebecca~A Betensky,
  Tim Clark, Sudeshna Das, and Bradley~T Hyman.
\newblock Tau induces blood vessel abnormalities and angiogenesis-related gene
  expression in p301l transgenic mice and human alzheimer’s disease.
\newblock {\em Proceedings of the National Academy of Sciences},
  115(6):E1289--E1298, 2018.

\bibitem{uginet2021cerebrovascular}
Marjolaine Uginet, Gautier Breville, J{\'e}r{\'e}my Hofmeister, Paolo Machi,
  Patrice~H Lalive, Andrea Rosi, Aikaterini Fitsiori, Maria~Isabel Vargas,
  Frederic Assal, Gilles Allali, et~al.
\newblock Cerebrovascular complications and vessel wall imaging in covid-19
  encephalopathy—a pilot study.
\newblock {\em Clinical neuroradiology}, pages 1--7, 2021.

\bibitem{blinder2013cortical}
Pablo Blinder, Philbert~S Tsai, John~P Kaufhold, Per~M Knutsen, Harry Suhl, and
  David Kleinfeld.
\newblock The cortical angiome: an interconnected vascular network with
  noncolumnar patterns of blood flow.
\newblock {\em Nature neuroscience}, 16(7):889--897, 2013.

\bibitem{todorov2020machine}
Mihail~Ivilinov Todorov, Johannes~Christian Paetzold, Oliver Schoppe, Giles
  Tetteh, Suprosanna Shit, Velizar Efremov, Katalin Todorov-V{\"o}lgyi, Marco
  D{\"u}ring, Martin Dichgans, Marie Piraud, et~al.
\newblock Machine learning analysis of whole mouse brain vasculature.
\newblock {\em Nature methods}, 17(4):442--449, 2020.

\bibitem{kirst2020mapping}
Christoph Kirst, Sophie Skriabine, Alba Vieites-Prado, Thomas Topilko, Paul
  Bertin, Gaspard Gerschenfeld, Florine Verny, Piotr Topilko, Nicolas
  Michalski, Marc Tessier-Lavigne, et~al.
\newblock Mapping the fine-scale organization and plasticity of the brain
  vasculature.
\newblock {\em Cell}, 180(4):780--795, 2020.

\bibitem{schmid2021severity}
Franca Schmid, Giulia Conti, Patrick Jenny, and Bruno Weber.
\newblock The severity of microstrokes depends on local vascular topology and
  baseline perfusion.
\newblock {\em Elife}, 10:e60208, 2021.

\bibitem{ueda2020tissue}
Hiroki~R Ueda, Ali Ert{\"u}rk, Kwanghun Chung, Viviana Gradinaru, Alain
  Ch{\'e}dotal, Pavel Tomancak, and Philipp~J Keller.
\newblock Tissue clearing and its applications in neuroscience.
\newblock {\em Nature Reviews Neuroscience}, 21(2):61--79, 2020.

\bibitem{erturk2012three}
Ali Ert{\"u}rk et~al.
\newblock Three-dimensional imaging of solvent-cleared organs using {3DISCO}.
\newblock {\em Nature Protocols}, 7(11):1983, 2012.

\bibitem{chung2013clarity}
Kwanghun Chung and Karl Deisseroth.
\newblock Clarity for mapping the nervous system.
\newblock {\em Nature methods}, 10(6):508--513, 2013.

\bibitem{renier2014idisco}
Nicolas Renier, Zhuhao Wu, David~J Simon, Jing Yang, Pablo Ariel, and Marc
  Tessier-Lavigne.
\newblock idisco: a simple, rapid method to immunolabel large tissue samples
  for volume imaging.
\newblock {\em Cell}, 159(4):896--910, 2014.

\bibitem{di2018whole}
Antonino~Paolo Di~Giovanna et~al.
\newblock Whole-brain vasculature reconstruction at the single capillary level.
\newblock {\em Scientific reports}, 8(1):12573, 2018.

\bibitem{kipf2016semi}
Thomas~N Kipf and Max Welling.
\newblock Semi-supervised classification with graph convolutional networks.
\newblock {\em arXiv preprint arXiv:1609.02907}, 2016.

\bibitem{kipf2016variational}
Thomas~N. Kipf and Max Welling.
\newblock Variational graph auto-encoders, 2016.

\bibitem{zeng2019graphsaint}
Hanqing Zeng, Hongkuan Zhou, Ajitesh Srivastava, Rajgopal Kannan, and Viktor
  Prasanna.
\newblock Graphsaint: Graph sampling based inductive learning method.
\newblock {\em arXiv preprint arXiv:1907.04931}, 2019.

\bibitem{hamilton2017inductive}
William~L Hamilton, Rex Ying, and Jure Leskovec.
\newblock Inductive representation learning on large graphs.
\newblock In {\em Proceedings of the 31st International Conference on Neural
  Information Processing Systems}, pages 1025--1035, 2017.

\bibitem{frasca2020sign}
Fabrizio Frasca, Emanuele Rossi, Davide Eynard, Ben Chamberlain, Michael
  Bronstein, and Federico Monti.
\newblock Sign: Scalable inception graph neural networks.
\newblock {\em arXiv preprint arXiv:2004.11198}, 2020.

\bibitem{klicpera2018predict}
Johannes Klicpera, Aleksandar Bojchevski, and Stephan G{\"u}nnemann.
\newblock Predict then propagate: Graph neural networks meet personalized
  pagerank.
\newblock {\em arXiv preprint arXiv:1810.05997}, 2018.

\bibitem{chiang2019cluster}
Wei-Lin Chiang, Xuanqing Liu, Si~Si, Yang Li, Samy Bengio, and Cho-Jui Hsieh.
\newblock Cluster-gcn: An efficient algorithm for training deep and large graph
  convolutional networks.
\newblock In {\em Proceedings of the 25th ACM SIGKDD International Conference
  on Knowledge Discovery \& Data Mining}, pages 257--266, 2019.

\bibitem{kong2020flag}
Kezhi Kong, Guohao Li, Mucong Ding, Zuxuan Wu, Chen Zhu, Bernard Ghanem, Gavin
  Taylor, and Tom Goldstein.
\newblock Flag: Adversarial data augmentation for graph neural networks, 2020.

\bibitem{huang2020combining}
Qian Huang, Horace He, Abhay Singh, Ser-Nam Lim, and Austin~R Benson.
\newblock Combining label propagation and simple models out-performs graph
  neural networks.
\newblock {\em arXiv preprint arXiv:2010.13993}, 2020.

\bibitem{klicpera2019directional}
Johannes Klicpera, Janek Gro{\ss}, and Stephan G{\"u}nnemann.
\newblock Directional message passing for molecular graphs.
\newblock In {\em International Conference on Learning Representations}, 2019.

\bibitem{zhang2021revisiting}
Muhan Zhang, Pan Li, Yinglong Xia, Kai Wang, and Long Jin.
\newblock Revisiting graph neural networks for link prediction, 2021.

\bibitem{zhang2018link}
Muhan Zhang and Yixin Chen.
\newblock Link prediction based on graph neural networks, 2018.

\bibitem{hu2020open}
Weihua Hu, Matthias Fey, Marinka Zitnik, Yuxiao Dong, Hongyu Ren, Bowen Liu,
  Michele Catasta, and Jure Leskovec.
\newblock Open graph benchmark: Datasets for machine learning on graphs.
\newblock {\em arXiv preprint arXiv:2005.00687}, 2020.

\bibitem{fey2019fast}
Matthias Fey and Jan~Eric Lenssen.
\newblock Fast graph representation learning with pytorch geometric.
\newblock {\em arXiv preprint arXiv:1903.02428}, 2019.

\bibitem{tetteh2018deepvesselnet}
Giles Tetteh et~al.
\newblock Deepvesselnet: Vessel segmentation, centerline prediction, and
  bifurcation detection in 3-d angiographic volumes.
\newblock {\em arXiv preprint arXiv:1803.09340}, 2018.

\bibitem{aanjaneya2011metric}
Mridul Aanjaneya, Frederic Chazal, Daniel Chen, Marc Glisse, Leonidas~J Guibas,
  and Dmitriy Morozov.
\newblock Metric graph reconstruction from noisy data.
\newblock In {\em Proceedings of the twenty-seventh annual symposium on
  Computational geometry}, pages 37--46, 2011.

\bibitem{meyer2009voreen}
Jennis Meyer-Spradow, Timo Ropinski, J{\"o}rg Mensmann, and Klaus Hinrichs.
\newblock Voreen: A rapid-prototyping environment for ray-casting-based volume
  visualizations.
\newblock {\em IEEE Computer Graphics and Applications}, 29(6):6--13, 2009.

\bibitem{drees2021scalable}
Dominik Drees, Aaron Scherzinger, Ren{\'e} H{\"a}gerling, Friedemann Kiefer,
  and Xiaoyi Jiang.
\newblock Scalable robust graph and feature extraction for arbitrary vessel
  networks in large volumetric datasets.
\newblock {\em arXiv preprint arXiv:2102.03444}, 2021.

\bibitem{schneider2012tissue}
Matthias Schneider et~al.
\newblock Tissue metabolism driven arterial tree generation.
\newblock {\em Med Image Anal.}, 16(7):1397--1414, 2012.

\bibitem{scherzinger2017interactive}
Aaron Scherzinger, Tobias Brix, Dominik Drees, Andreas V{\"o}lker, Kiril
  Radkov, Niko Santalidis, Alexander Fieguth, and Klaus~H Hinrichs.
\newblock Interactive exploration of cosmological dark-matter simulation data.
\newblock {\em IEEE computer graphics and applications}, 37(2):80--89, 2017.

\bibitem{brix2014visualization}
Tobias Brix, J{\"o}rg-Stefan Pra{\ss}ni, and Klaus Hinrichs.
\newblock Visualization of large volumetric multi-channel microscopy data
  streams on standard pcs.
\newblock {\em arXiv preprint arXiv:1407.2074}, 2014.

\bibitem{dierkes2018three}
Cathrin Dierkes, Aaron Scherzinger, and Friedemann Kiefer.
\newblock Three-dimensional visualization of the lymphatic vasculature.
\newblock In {\em Lymphangiogenesis}, pages 1--18. Springer, 2018.

\bibitem{hagerling2017vipar}
Ren{\'e} H{\"a}gerling, Dominik Drees, Aaron Scherzinger, Cathrin Dierkes,
  Silvia Martin-Almedina, Stefan Butz, Kristiana Gordon, Michael Sch{\"a}fers,
  Klaus Hinrichs, Pia Ostergaard, et~al.
\newblock Vipar, a quantitative approach to 3d histopathology applied to
  lymphatic malformations.
\newblock {\em JCI insight}, 2(16), 2017.

\bibitem{bovenkamp2015velocity}
Philipp~Rene Bovenkamp, Tobias Brix, Florian Lindemann, Richard Holtmeier,
  Desiree Abdurrachim, Michael~T Kuhlmann, Gustav~J Strijkers, J{\"o}rg
  Stypmann, Klaus~H Hinrichs, and Verena Hoerr.
\newblock Velocity mapping of the aortic flow at 9.4 t in healthy mice and mice
  with induced heart failure using time-resolved three-dimensional
  phase-contrast mri (4d pc mri).
\newblock {\em Magnetic Resonance Materials in Physics, Biology and Medicine},
  28(4):315--327, 2015.

\bibitem{lee1994building}
Ta-Chih Lee et~al.
\newblock Building skeleton models via {3-D} medial surface axis thinning
  algorithms.
\newblock {\em CVGIP: Graphical Models and Image Processing}, 56(6):462--478,
  1994.

\bibitem{isenburg2009streaming}
Martin Isenburg and Jonathan Shewchuk.
\newblock Streaming connected component computation for trillion voxel images.
\newblock In {\em Workshop on Massive Data Algorithmics}, volume~2, 2009.

\bibitem{sunkin2012allen}
Susan~M Sunkin, Lydia Ng, Chris Lau, Tim Dolbeare, Terri~L Gilbert, Carol~L
  Thompson, Michael Hawrylycz, and Chinh Dang.
\newblock Allen brain atlas: an integrated spatio-temporal portal for exploring
  the central nervous system.
\newblock {\em Nucleic acids research}, 41(D1):D996--D1008, 2012.

\bibitem{hahn2021large}
Artur Hahn, Julia Bode, Allen Alexander, Kianush Karimian-Jazi, Katharina
  Schregel, Daniel Schwarz, Alexander~C Sommerkamp, Thomas Kr{\"u}wel, Amir
  Abdollahi, Wolfgang Wick, et~al.
\newblock Large-scale characterization of the microvascular geometry in
  development and disease by tissue clearing and quantitative ultramicroscopy.
\newblock {\em Journal of Cerebral Blood Flow \& Metabolism}, 41(7):1536--1546,
  2021.

\bibitem{gross2018graph}
Jonathan~L Gross, Jay Yellen, and Mark Anderson.
\newblock {\em Graph theory and its applications}.
\newblock Chapman and Hall/CRC, 2018.

\bibitem{grover2016node2vec}
Aditya Grover and Jure Leskovec.
\newblock node2vec: Scalable feature learning for networks.
\newblock In {\em Proceedings of the 22nd ACM SIGKDD international conference
  on Knowledge discovery and data mining}, pages 855--864, 2016.

\bibitem{katz1953new}
Leo Katz.
\newblock A new status index derived from sociometric analysis.
\newblock {\em Psychometrika}, 18(1):39--43, 1953.

\bibitem{adamic2003friends}
Lada~A Adamic and Eytan Adar.
\newblock Friends and neighbors on the web.
\newblock {\em Social networks}, 25(3):211--230, 2003.

\bibitem{schlichtkrull2018modelingrgcn}
Michael Schlichtkrull, Thomas~N Kipf, Peter Bloem, Rianne Van Den~Berg, Ivan
  Titov, and Max Welling.
\newblock Modeling relational data with graph convolutional networks.
\newblock In {\em European semantic web conference}, pages 593--607. Springer,
  2018.

\bibitem{vashishth2019composition}
Shikhar Vashishth, Soumya Sanyal, Vikram Nitin, and Partha Talukdar.
\newblock Composition-based multi-relational graph convolutional networks.
\newblock {\em arXiv preprint arXiv:1911.03082}, 2019.

\bibitem{dyer2017quantifying}
Eva~L Dyer, William~Gray Roncal, Judy~A Prasad, Hugo~L Fernandes, Doga
  G{\"u}rsoy, Vincent De~Andrade, Kamel Fezzaa, Xianghui Xiao, Joshua~T
  Vogelstein, Chris Jacobsen, et~al.
\newblock Quantifying mesoscale neuroanatomy using x-ray microtomography.
\newblock {\em Eneuro}, 4(5), 2017.

\bibitem{wu20143d}
Jingpeng Wu, Yong He, Zhongqin Yang, Congdi Guo, Qingming Luo, Wei Zhou,
  Shangbin Chen, Anan Li, Benyi Xiong, Tao Jiang, et~al.
\newblock 3d braincv: simultaneous visualization and analysis of cells and
  capillaries in a whole mouse brain with one-micron voxel resolution.
\newblock {\em Neuroimage}, 87:199--208, 2014.

\bibitem{calabrese2015diffusion}
Evan Calabrese, Alexandra Badea, Gary Cofer, Yi~Qi, and G~Allan Johnson.
\newblock A diffusion mri tractography connectome of the mouse brain and
  comparison with neuronal tracer data.
\newblock {\em Cerebral cortex}, 25(11):4628--4637, 2015.

\bibitem{li2019contrast}
Tianqi Li, Chao~J Liu, and Taner Akkin.
\newblock Contrast-enhanced serial optical coherence scanner with deep learning
  network reveals vasculature and white matter organization of mouse brain.
\newblock {\em Neurophotonics}, 6(3):035004, 2019.

\bibitem{miettinen2021micrometer}
Arttu Miettinen, Antonio Zippo, Alessandra Patera, Anne Bonnin, Sarah
  Shahmoradian, Gabriele Biella, and Marco Stampanoni.
\newblock Micrometer-resolution reconstruction and analysis of whole mouse
  brain vasculature by synchrotron-based phase-contrast tomographic microscopy.
\newblock {\em bioRxiv}, 2021.

\bibitem{kullback1951information}
Solomon Kullback and Richard~A Leibler.
\newblock On information and sufficiency.
\newblock {\em The annals of mathematical statistics}, 22(1):79--86, 1951.

\bibitem{ying2019gnnexplainer}
Rex Ying, Dylan Bourgeois, Jiaxuan You, Marinka Zitnik, and Jure Leskovec.
\newblock Gnnexplainer: Generating explanations for graph neural networks.
\newblock {\em Advances in neural information processing systems}, 32:9240,
  2019.

\end{thebibliography}
}
\appendix

\textbf{Supplementary material}
\section{Additional graph visualisations}
\begin{figure}[ht!]
    \centering

    \includegraphics[width=1\textwidth]{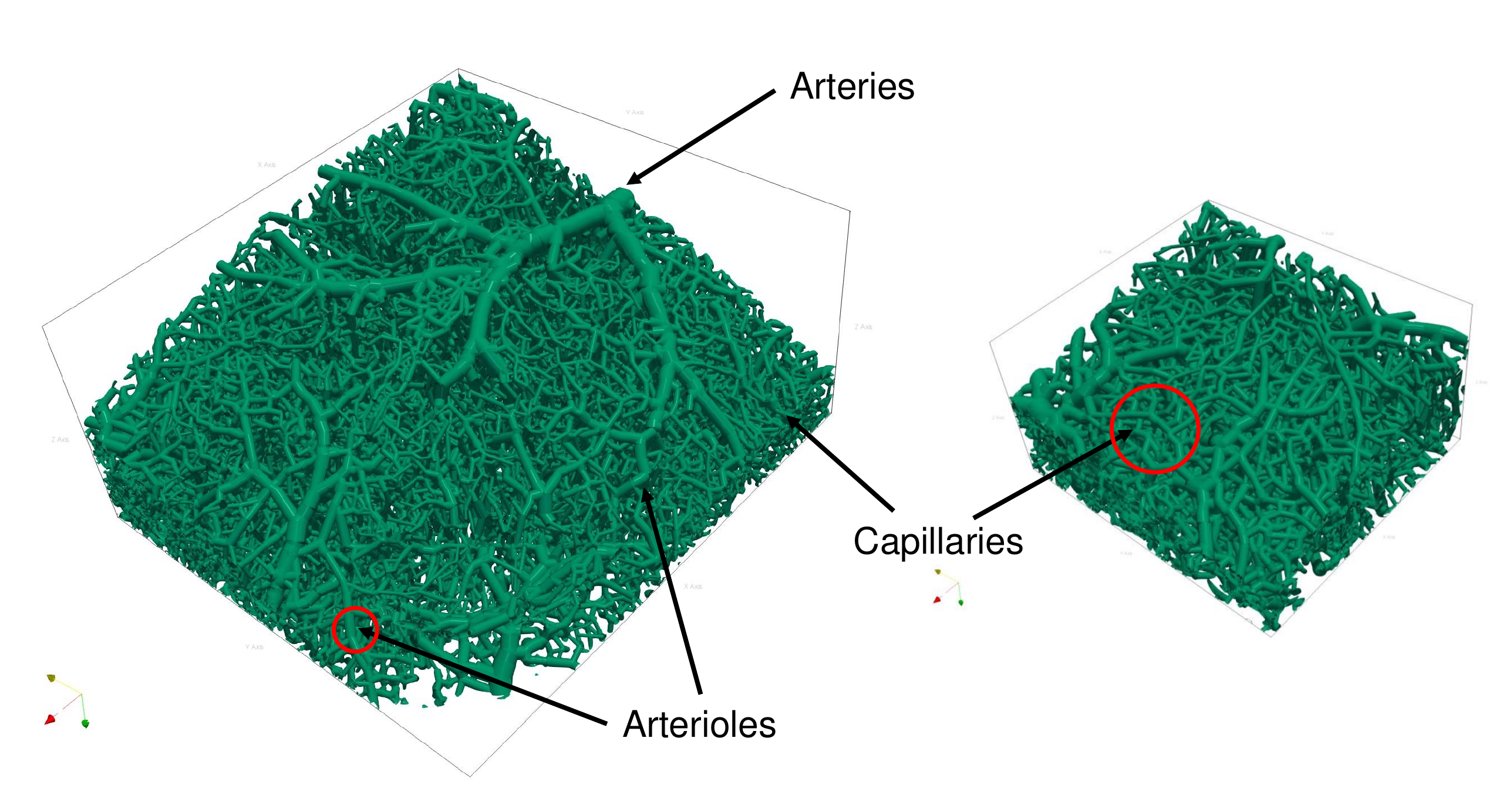}
    \caption{Graphical visualisations of the vessel graph and three different vessel types distinguishable by diameter.}
    \label{fig:suppl_vessel_types}
\end{figure}
\section{Code and dataset documentation}

\subsection{Hosting, licensing and author statement }
\label{Licensing}

All of our codes and baselines are available in a public github repository \url{https://github.com/jocpae/VesselGraph} licensed under an MIT License. All of our data is also freely available and can be downloaded from a university server following the links in \url{https://github.com/jocpae/VesselGraph#datasets}. The dataset is provided in CSV format. The dataset has the following DOI: \url{10.5281/zenodo.5301621}. All of our released data is licensed under an \textit{Attribution-NonCommercial 4.0 International (CC BY-NC 4.0) license}. The authors confirm the CC licenses for the included datasets and declare to bear all responsibility in case of violation of rights. The authors declare no competing financial interests

\subsection{Long term maintenance plan}
The dataset and code has been permanently archived at Zenodo, guaranteeing long-term availability. We will update the dataset when novel segmentations of the whole brain vasculature become publicly available. Contributions will be solicited via GitHub pull request. Regarding maintenance, we will update the code repository for loading and processing the data; the links to the university server where the data is stored will also be kept up-to-date.

Additionally, we plan to incorporate our dataset into the Open Graph Benchmark\footnote{\url{https://ogb.stanford.edu/}}. Our dataset and dataloader are already compatible with the OGB data loader and platform. We believe that an integration into the OGB framework will further facilitate and simplify the usage of our data. We also provide an alternative dataloader for use with Pytorch Geometric.

\subsection{Whole brain vessel imaging and segmentation}
\label{Whole_Brain_Imaging_Techniques}

\begin{figure}[ht]
    \centering
    \includegraphics[width=1\textwidth]{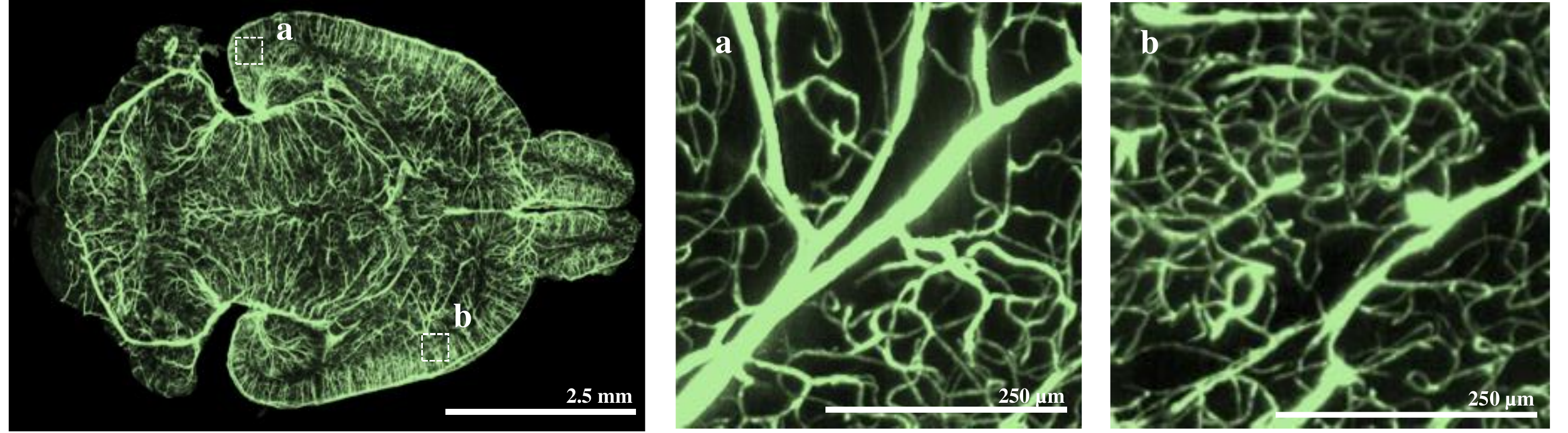}
    \caption{Local differences in vessel structure can be observed in different brain regions in a dataset of Todorov and Paetzold et al. \cite{todorov2020machine}.}
    \label{fig:vessel_regions}
\end{figure}

As discussed in the introduction, whole brain vascular imaging is an emerging field spanning different imaging techniques, among the first technologies were tissue clearing-based methods \cite{ueda2020tissue,erturk2012three,chung2013clarity,renier2014idisco} which enable the fluorescent staining and clearing (that is, a chemical process which renders the organ transparent) of intact, whole brains with a subsequent imaging of the vessels with a 3D lightsheet microscope. Among the first imaging protocols were VesSAP \cite{todorov2020machine}, Tubemap \cite{kirst2020mapping} and the work by diGiovanna et al. \cite{di2018whole}. Alternative imaging techniques are microCT\cite{dyer2017quantifying,wu20143d}, magnetic resonance imaging \cite{calabrese2015diffusion} and optical coherence tomography \cite{li2019contrast}, which however fail to achieve the spatial resolution to reliably image microcapillaries. Recently, a method based on synchrotron-based phase-contrast tomographic microscopy was developed, achieving an isotropic voxel size of $0.65$ micrometers \cite{miettinen2021micrometer}. Other technologies, such as serial serial two-photon microscopic imaging are also developing rapidly with similar or even better resolution compared to the tissue clearing methods (e.g. $0.303 \times 0.303 \times 1.0$ resolution \cite{ji2021brain}), promising a widespread use/adoption of whole brain vascular imaging approaches in the future.

\subsection{Individual datasets, licenses and animal experiments}
\label{Individual_Licenses_Data}
The nine base datasets from the VesSAP paper \cite{todorov2020machine} are available here: \url{http://discotechnologies.org/VesSAP/}. They are licensed under a Attribution-NonCommercial 4.0 International (CC BY-NC 4.0). The animal experiments were carried out under approval of the institutional ethics review board of the Government of Upper Bavaria (Regierung von Oberbayern, Munich, Germany), and in accordance with European directive 2010/63/EU for animal research, details can be read here \url{https://doi.org/10.1038/s41592-020-0792-1}.

The base datasets from Ji et al. \cite{ji2021brain} are licensed under an open source (BSD 3-Clause) license. They are available upon email request from the authors, their code is available at \url{https://neurophysics.ucsd.edu/software.php}. The animal experiments followed the Guide for the Care and Use of Laboratory Animals and have been approved by the Institutional Animal Care and Use Committee, details can be found in the original paper: \url{https://doi.org/10.1016/j.neuron.2021.02.006}

The synthetic data was generated by the authors themselves following the approach by Schneider et al. \cite{schneider2012tissue}, the same license applies as for the graph datasets presented here. The synthetic base data can be downloaded here \url{https://github.com/giesekow/deepvesselnet/wiki/Datasets}. 


\clearpage
\section{Graph documentation}
\label{Suppl_Graph_Doc}

\begin{figure}[htp!]
    \centering
    \includegraphics[width=1.0\textwidth]{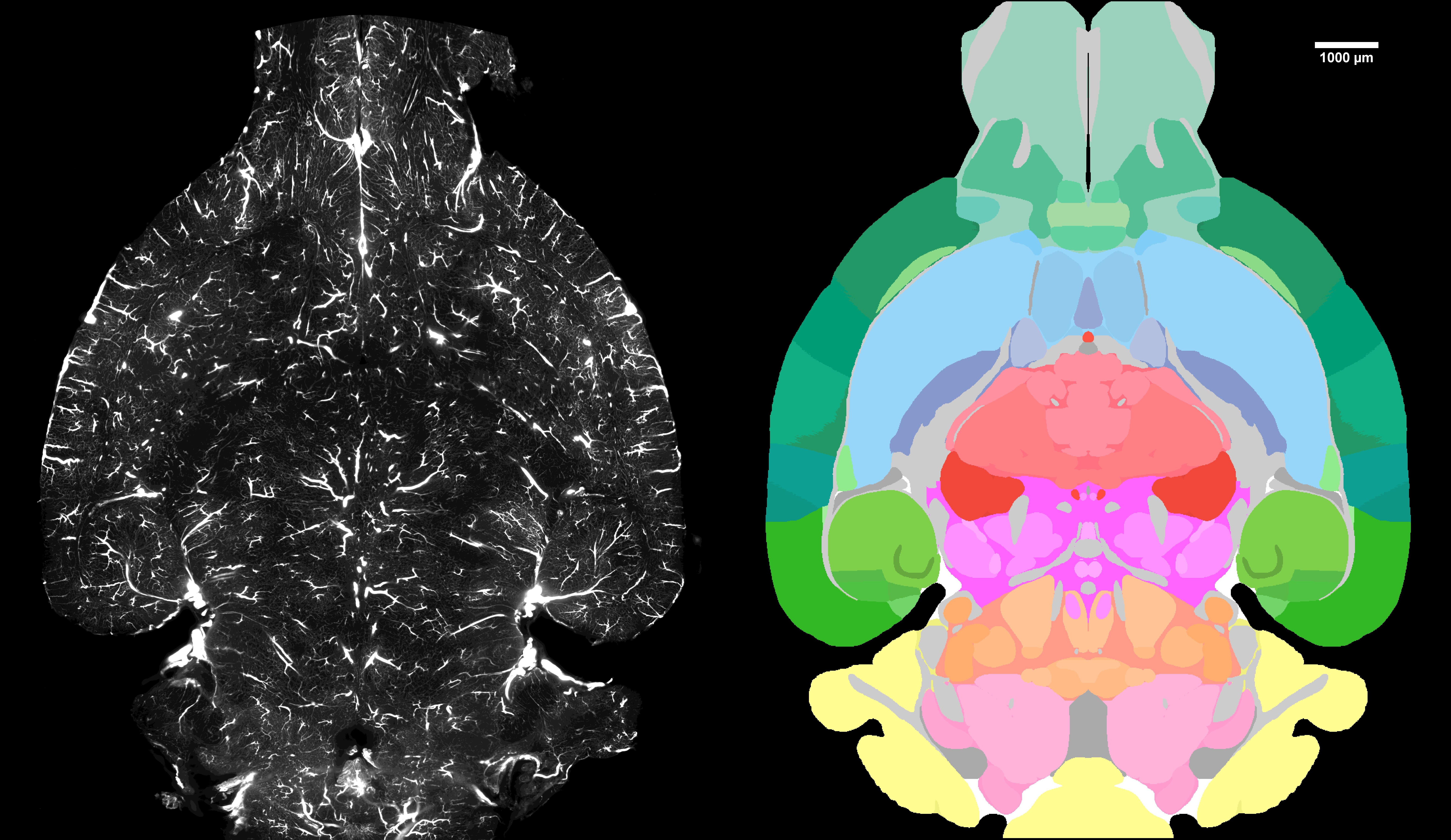}
    \caption{Rendering of the Allen brain atlas feature; on the left side a 2D image slice from the CD1-E-1 brain is depicted, on the right side a rendering of the Allen brain atlas regions corresponding to the coordinates of the image. A node $n$ is assigned to a particular atlas region depending on the ${x_n,y_n,z_n}$ coordinate of the node $n$ . }
    \label{fig:atlas_image}
\end{figure}

\subsection{Voreen parameters}
\label{Suppl_Graph_Extraction}
In the following Sections, we discuss the details of the Voreen graph extraction pipeline which follows a four stage protocol introduced in Section \ref{Graph_Ext}. To recapitulate, the four stages are: 

\begin{enumerate}
    \item Skeletonization: The binary segmentation volume is reduced to a skeleton based representation by applying a standard topological thinning algorithm by Lee et al. \cite{lee1994building}.
    \item Topology Extraction: memory efficient algorithms extract the vessel centerlines \cite{isenburg2009streaming}. \textit{Voreen} allows to store this intermediate representation in a combination with the graph. 
    \item Voxel-Branch Assignment: Computing of mapping between the so-called protograph (i.e. the initial graph) and the voxels of the binary segmentation.
    \item Feature Extraction: On basis of the protograph and the mapping, several features can be computed from the foreground segmentation.
\end{enumerate}

We chose the \textit{Voreen} parameters in the following manner: 

\begin{enumerate}
    \item The \textit{binarization threshold} is selected from the interval $[0,1]$. This value is irrelevant for binary segmentation masks, e.g. VesSAP.
    \item  \textit{Surface smoothing} is deactivated.
    \item The relative \textit{minimal bounding box diagonal} is set to $0.05$.
    \item  The total \textit{minimal bounding box diagonal} is set to $0$ mm.
    \item The \textit{bulge size} is set to $3.0$, see Figure \ref{fig:bulge}.
\end{enumerate}
\paragraph{Bulge Size}

As depicted in the figures below, the total number of nodes decreases when increasing the bulge size parameter. This is expected, as the bulge size describes the relation between parent vessel and branch, and the relation between vessel \textit{bumpiness} and parent vessel \cite{drees2021scalable}.
The bulge size can be configured between $[0 <x< 10]$. \\

\begin{figure}[ht]
    \centering
    \includegraphics[width=1.0\textwidth]{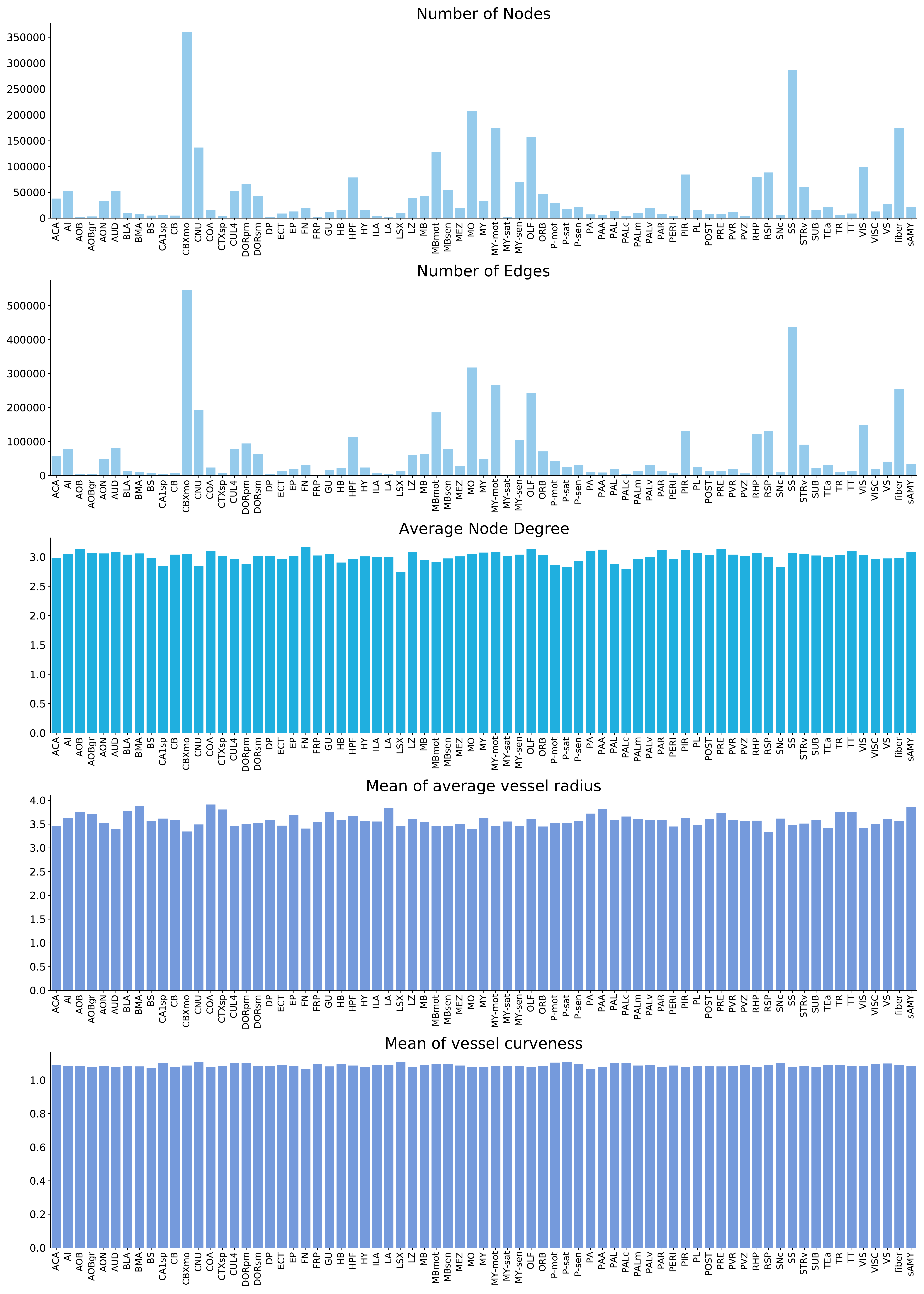}
    \caption{Different Node features of $G$  plotted by the regions of the Allen brain atlas.}
    \label{fig:feats_by_region}
\end{figure}

\clearpage

\begin{figure}[ht]
    \centering
    \includegraphics[width=1.0\textwidth]{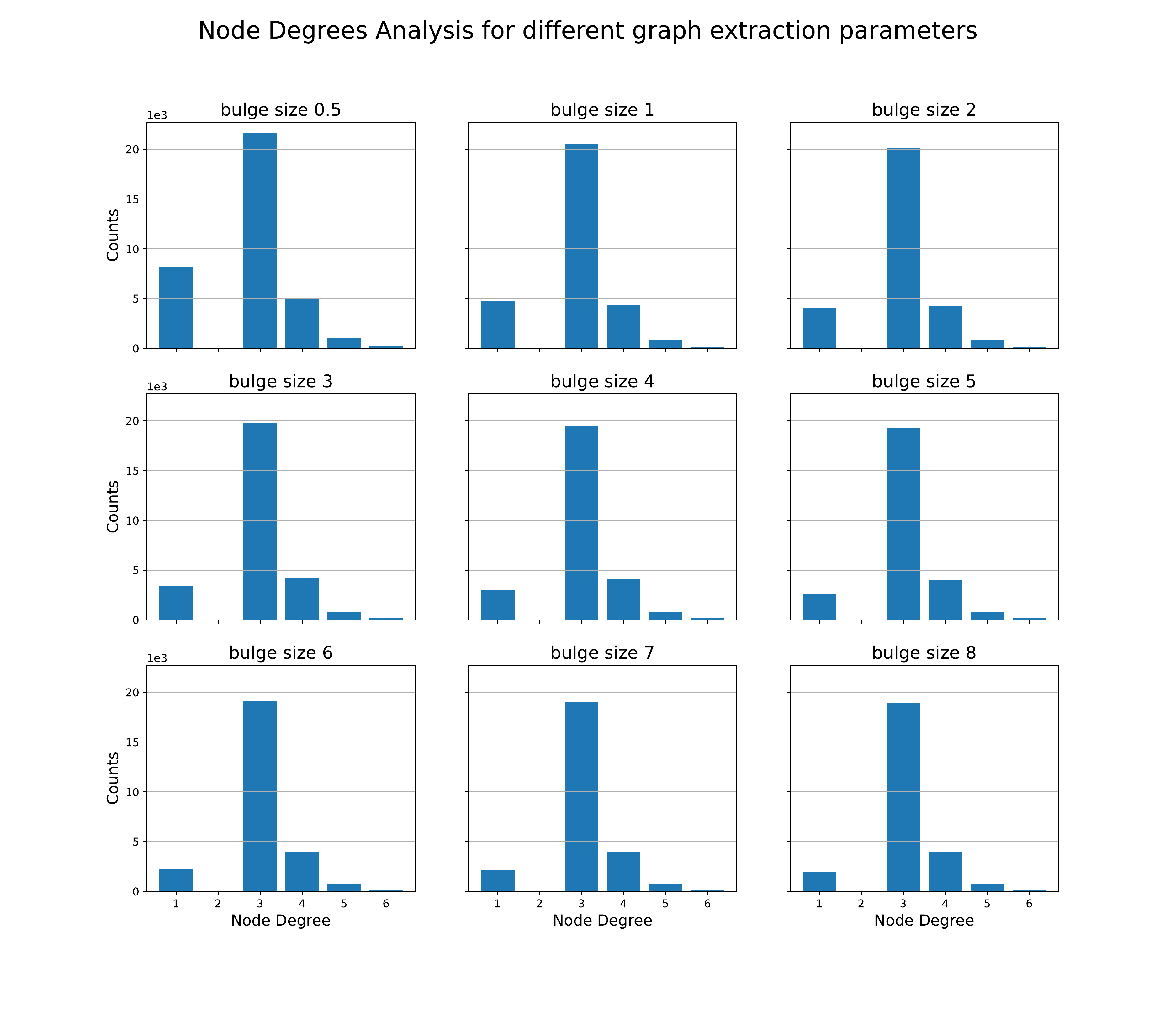}
    \caption{Bulge Size analysis in the \textit{Voreen} pipeline for a parameter range of 0.5- 8.0.}
    \label{fig:bulge}
\end{figure}

For bulge \textunderscore{size} $= 3.0$, we obtain a good relationship between capturing the essential vessels, while capturing the smoothness of the vascular trees. Following the recommendation of \textunderscore{size} $= 3.0$ \cite{drees2021scalable} for healthy vasculature, we keep \textunderscore{size} $= 3.0$ as a maximum, to provide a reasonable \textunderscore{size} $= 3.0$ value for future data with diseased animals.\\

\begin{figure}[htp!]
    \centering
    \includegraphics[width=\textwidth]{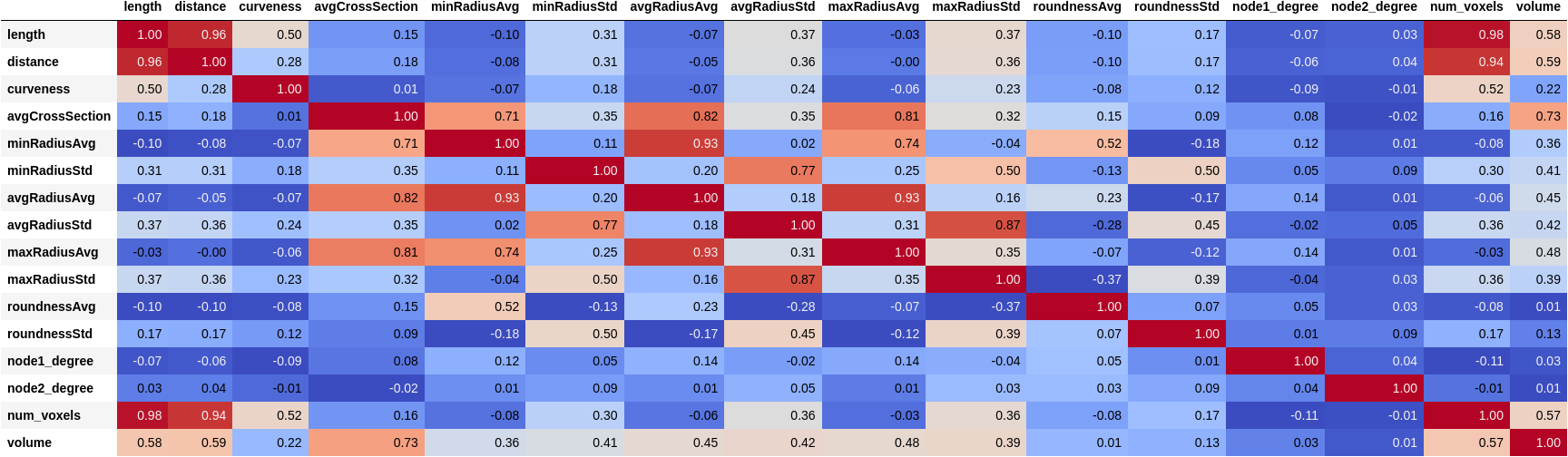}
    \caption{Edge feature correlation plot.}
    \label{fig:edge_corr}
\end{figure}

\paragraph{Runtime of graph extraction:}
The \textit{Voreen} pipeline used for graph extraction was run on a CPU cluster. We timed one exemplary graph extraction on this cluster using 8 logical CPUs. For one of our synthetic vessel datasets ($500 \times 500 \times 401$ pixels), the process described in Section \ref{Graph_Ext} with $3$ refinement iterations, resulting in $28538$ nodes and $42727$ edges required a total runtime of $3$ minutes and $57$ seconds. Importantly, the extraction times scale roughly linearly with the number of nodes which are extracted \cite{drees2021scalable}.

\subsection{Post processing of graphs}
\label{Suppl_Post_Processing}

In order to remove evident artifacts from the generated graphs we implemented rule based post processing or pruning steps. 

\paragraph{Feature merging:}
In a small percentage of the extracted graphs ($<1\%$), we obtain multiple edges $e^{b}_{ij}$ for $b=1:B$ with the same vertices. This can be attributed to imaging artifacts and irregularities in the vessel staining process. For instance, holes in the segmentation mask, if not properly filled, can result in multiple edges spanning from the same source node $n_{i}$ to the same target node $n_{j}$, respectively $n_{j}$ to $n_{i}$, as the underlying graph is undirected. As our focus lies on biologically realistic graphs, in particular correct branching structures, we merge the edge features of two identically labeled edges by obtain approximations. As the greater percentage of irregularities have been already mitigated in various preprocessing steps, we are confident that approximations in a small number of edges should not affect the performance and generalization of the deep learning models. When we merge features we update them according to:

\begin{enumerate}
    \item Length: $l_{ij}  = \frac{1}{B} \sum_{b=1}^{B} {l_{ij}^{b}} $ 
    \item Shortest Distance: $d_{ij} = \min_k\{d_{ij}^{b}\}$
    \item Volume: $v_{ij}  = \sum_{b=1}^{B} {v_{ij}^{b}} $ 
    \item Number of Voxels: $nv_{ij} = \sum_{b=1}^{B} {nv_{ij}^{b}} $ 
    \item Curvature: $\rho_{ij}  = \frac{1}{B} \sum_{b=1}^{B} {\rho_{ij}^{b}} $ 
    \item Mean Cross Section Area: $\alpha_{ij}  = \sum_{b=1}^{B} {\alpha_{ij}^{b}} $ 
\end{enumerate}

As radius features are highly shape-dependent, and highly variable in the vessel itself, we can only approximate the vessel attributes, as we cannot access the meta-information Voreen utilizes to calculate the minimum, average and maximum vessel radius and the corresponding standard deviations. 

Consequently, we obtain the average radius by summing the average radius of all identically labeled edges. We calculate the standard deviation by obtaining the median of the mathematical relationship of standard deviation of average radius to average radius. 

\begin{enumerate}
    \item  $\mu_{ij}^{r} = \mu_{ij}^{\bar{r}} = \mu_{ij}^{R} = \sum_{b=1}^{B} \mu_{ij}^{\bar{r}b}$
    \item $\sigma _{ij}^{r} =\sigma _{ij}^{\bar{r}} = \sigma _{ij}^{R} =\sqrt{\sum_{b=1}^{B}(\sigma _{ij}^{\bar{r}b})^2}$
\end{enumerate}

We fit a linear function $\mu^{o} =f(\mu^{\overline{r}}) = a \cdot \mu^{\overline{r}}+b$ to map the average radius $\mu^{\overline{r}}$ to the vessel roundness $\mu^{o}$. Similarly to the radius approximations, we obtain the newly computed roundness standard deviation by obtaining the median of the mathematical relationship of standard deviation to roundness.

\begin{enumerate}
    \item $\mu_{ij}^{o} =f(\mu_{ij}^{\overline{r}})$
    \item $\sigma _{ij}^{o} = median\left[ \left\{\frac{\sigma_{ij}^{ob}}{\mu_{ij}^{ob}}\right\}_{b=1:B}\right] \cdot \mu_{ij}^{o}$
\end{enumerate}

The following properties are computationally updated.
\begin{itemize}
    \item $\nu_{ij}^{1}$ is updated. 
    \item $\nu_{ij}^{2}$ is updated. 
\end{itemize}

\section{Baseline experimentation and discussion}
\subsection{Link prediction}
\label{Suppl_Link_Pred}
\begin{table}[h]
\centering
\tiny
\caption{\label{tab:link_hyper_parameters} Details of the hyper-parameters search for Link prediction with the selected parameters for the final training of each of our baseline models.}

\begin{tabular}{| l||l|l|l|}
\hline

Model                                    & Parameter Range & Selected Parameters & Model Select.\\
\hline
Matrix Factorization                 & lr $\in$ \{$1 \cdot10^{-3}, 1\cdot 10^{-4},
                                                1 \cdot10^{-5}$\}
                                                & lr = $1 \cdot10^{-5}$ &  epochs $3000$  \\
                                                & num of layers  $\in$ \{$2,3,4$\}          
                                                & num of layers  = $3$ &  \\
                                                & hidden channels  $\in$ \{$32,64,128,256$\}      
                                                & hidden channels = $64$ &  \\
                                                & dropout  $\in$ \{$0,0.2,0.5$\}
                                                & dropout = $0.2$ & \\

\hline 
MLP                                             & lr $\in$ \{$1 \cdot10^{-2},1 \cdot10^{-3},                                                    1\cdot 10^{-4}, 1 \cdot10^{-5}$\}
                                                & lr = $1 \cdot10^{-5}$ &  epochs $3000$  \\
                                                & num of layers  $\in$ \{$2,3,4$\}          
                                                & num of layers  = $4$  &  \\
                                                & hidden channels  $\in$ \{$128,256,512$\}      
                                                & hidden channels = $256$ &  \\
                                                & dropout  $\in$ \{$0,0.2,0.5$\}
                                                & dropout = $0.2$ & \\

\hline 
GCN    \cite{kipf2016semi}           & lr $\in$ \{$1 \cdot10^{-2},1 \cdot10^{-3}, 1\cdot 10^{-4},                                                      1\cdot 10^{-5}$\}
                                                & lr = $1 \cdot10^{5}$ &  epochs $3000$  \\
                                                & num of layers  $\in$ \{$2,3,4$\}          
                                                & num of layers  = $2$  &  \\
                                                & hidden channels  $\in$ \{$128,256,512$\}      
                                                & hidden channels = $256$ &  \\
                                                & dropout  $\in$ \{$0,0.2,0.5$\}
                                                & dropout = $0.2$ & \\
\hline 
GNN + N2Vec Embeddings \cite{grover2016node2vec}                                            &                                                  lr $\in$ \{$1 \cdot10^{-2},1 \cdot10^{-3}, 
                                                1\cdot 10^{-4}, 1\cdot 10^{-5}$\}
                                                & lr = $1 \cdot10^{-5}$ &  epochs $3000$  \\
                                                & num of layers  $\in$ \{$2,3,4$\}          
                                                & num of layers  = $2$  &  \\
                                                & hidden channels  $\in$ \{$128,256,512$\}      
                                                & hidden channels = $256$ &  \\
                                                & dropout  $\in$ \{$0,0.2,0.5$\}
                                                & dropout = $0.2$ & \\
\hline 
GNN + SAGE \cite{hamilton2017inductive}  
                                                & lr $\in$ \{$1 \cdot10^{-2},1 \cdot10^{-3},
                                                1\cdot 10^{-4},1\cdot 10^{-5}$\}
                                                & lr = $1 \cdot10^{-4}$ &  epochs $3000$  \\
                                                & num of layers  $\in$ \{$2,3,4$\}          
                                                & num of layers  = $3$  &  \\
                                                & hidden channels  $\in$ \{$128,256,512$\}      
                                                & hidden channels = $16$ &  \\
                                                & dropout  $\in$ \{$0,0.2,0.5$\}
                                                & dropout = $0.5$ & \\

\hline 
GNN + SAGE \cite{hamilton2017inductive}  + N2Vec  Embeddings \cite{grover2016node2vec}                                                  & lr $\in$ \{$1 \cdot10^{-02},1 \cdot10^{-3},
                                                1\cdot 10^{-4},1\cdot 10^{-5}$\} 
                                                & lr = $1 \cdot10^{-3}$ &  epochs $3000$  \\
                                                & num of layers  $\in$ \{$2,3,4$\}          
                                                & num of layers  = $2$  &  \\
                                                & hidden channels  $\in$ \{$128,256,512$\}      
                                                & hidden channels = $16$ &  \\
                                                & dropout  $\in$ \{$0,0.2,0.5$\}
                                                & dropout = $0.5$ & \\
\hline 
SEAL \cite{zhang2018link}
                                                & lr $\in$ \{$1\cdot 10^{-4},1\cdot 10^{-5}$\}
                                                & lr = $ 1 \cdot10^{-4}$\} &  epochs $100$  \\
                                                & num of layers  $\in$ \{$2,3,4$\}          
                                                & num of layers  = $2$  &  \\
                                                & hidden channels  $\in$ \{$32,64,128$\}      
                                                & hidden channels = $32$ &  \\
                                                & dropout  $\in$ \{$0,0.2,0.5$\}
                                                & dropout = $0.0$ & \\
                                                & num of hops$\in$ \{$1,2,3$\}      
                                                & num of hops  = $1$ &  \\
                                                & labeling $\in$ \{drnl,de,de+,zo\}      
                                                & labeling = drnl & \\
                                                & model = {DGCNN, GCN}
                                                & model = {DGCNN} &\\
\hline
N2Vec \cite{grover2016node2vec}              & lr $\in$ \{$1\cdot10^{-2}, 1\cdot 10^{-3},
                                                              1 \cdot10^{-4},1 \cdot10^{-5},1 \cdot10^{-6}$\}
                                                & lr = $1 \cdot10^{-6}$\ &  epochs $2$  \\
                                                & walk length $\in$ \{$5,10,20$\}          
                                                & walk length = $5$  & \\
                                                & walks per\_node $\in$ \{$5,10,20$\}     
                                                & walks\_per\_node  = $10$ &  \\
                                                & embedding\_dim $\in$ \{$16,32,64,128,256$\}
                                                & embedding\_dim = $64$& \\
\hline                                                
\end{tabular}

\end{table}

We implemented a set of baselines from the literature. All of these implementations, including documentation are available in our GitHub repository \url{}. The model architecture, learning rate, number of GCN layers, dropout and batch size  are the hyperparameters we optimized for our training and document below. We selected our models according to the highest ROC AUC score on the validation set. 

In order to select the values, we employed Grid Search, please see the tables below. Owing to the huge size of our dataset, hyperparameter tuning is challenging. To overcome this challenge, we subsample a region of the mouse brain in order to create a small graph. To ensure we are not introducing any bias, we measured the KL-Divergence \cite{kullback1951information} to ensure that our small graph is representative of the whole brain in its distribution of vasculature. We selected the best set of hyperparameters on the small graph and used it on the actual graph with small modifications if needed. We summarize our results in Table \ref{tab:link_hyper_parameters} and Table \ref{tab:node_hyper_parameters}.

For Matrix Factorization, we tune the number of layers, hidden channels and dropout rate. To estimate the best hyperparameters, we employed grid search. We observe that the Matrix Factorization method is not sensitive to the choice of hyper-parameters as long as they are in a reasonable range (learning rate 1e-3 to 1e-5, dropout 0 to 0.5, hidden channels 32 to 256). We obtained the best results for a learning rate of 1e-4 with 3 layers, 64 hidden channels and a dropout rate of 0.2.  
When extending to the entire BALBc1 whole mouse brain dataset (5.35 million edges, 3.54 million nodes), we reduced the number of hidden channels to 64 owing to GPU memory constraints. We experimentally found out that the model with 64 channels was easily able to overfit to the training data, showing that the model is sufficiently complex. 

The MLP model is trained with the Adam optimizer and a learning rate of 1e-4. This combination of hyperparameters provides us the best performance on the validation and test splits. We experimented with fully connected layers of 128 and 256 features, the latter resulting in better scores. Thus, our final model has 4 layers each with 256 channels.

For the Graph Neural Networks, we use two setups. A GCN setup without embeddings (which we refer to as GCN in the experiments) and a GCN with embeddings (which we refer to as GCN + Node2Vec Embeddings in the experiments, Tab. \ref{tab:link_hyper_parameters}). Both models are trained with Adam and 1e-5 and 1e-7 learning rate, respectively.

We discover that GCN+node embeddings do not perform superior to the GCN which is trained only on the node features. Moreover, in our experiments, we find that the predictions made by the models result in scores close to chance (approx. 51 for both). The model is overfitting on the training data almost immediately and is not able to generalize to the unseen data. Reducing the model complexity by reducing the number of layers, hidden channels and increasing the dropout did not improve performance.

We observe poor generalization and immediate overfitting in all cases. We hypothesize that this behaviour is caused by the model not being able to distinguish symmetrically placed nodes while making the prediction (something that the SEAL algorithm achieves using the Labeling trick). Our experiments done using SEAL with a similar set of hyper-parameters substantiate this hypothesis. Please refer to the section on SEAL for more details below.

In our experiments, we find that the GraphSAGE models tend to outperform the GCN baseline. The \textit{ROC AUC} increases to approximately $60$ meaning a performance boost of almost $7$\% We initialized the nodes with the Node2Vec embeddings\cite{grover2016node2vec} (we refer to this model as GraphSAGE + Node2Vec Embeddings) and findthat the inclusion of pretrained Node2Vec embeddings as features leads to deterioration in model performance. We observed the best model performance with a shallow network (2 hidden layers), a learning rate of $1e-4$ and $0.5$ dropout. The hyperparameters were chosen according to the GCN baseline.

When we ran our model on the entire mouse brain graph, we observed behaviour similar to GCN and GCN+embedding. Although the GraphSAGE performance is slightly superior to the GCN baseline, it is still dwindling in the region of being random (approx 59\%). The GraphSAGE algorithm tends to improve the update of node features by concatenating the features of the node and its aggregated neighbourhood features. As such, this may enable the network to assign different weights to the self nodes and its neighbours. However, this does not allow it to break the symmetry while performing the link prediction task. Indicating that these models suffer from the same issue in regards to node labeling as was described in the GCN discussion above. Further, inclusion of Node2Vec features decreases the performance resulting in an $53\%$ ROC AUC value.

Since Node2Vec is an unsupervised process we use the SGD optimizer with negative sampling, to maximize the log likelihood objective on the random walks in the graph. We perform a hyperparameter search on the learning rate, walk lengths, walks per node and embedding dimension.

We find the best combination to be a walk length of 5, 10 walks per node and an embedding dimension of 64. In our experiments with the small sub-graph, we observed that the loss quickly plateaus after 3 epochs. We use the same hyperparameter combination for the entire mouse brain graph. We observe that the inclusion of Node2Vec features does not improve performance and in most cases, leads to worse results. Empirically we observe that the inclusion of Node2Vec features is not useful for the link prediction task and as discussed above, the Labeling trick constitutes a much more important component.

\paragraph{SEAL:} For the SEAL algorithm we implement a similar hyperparameter search. We discover a general pattern in the selection of different labeling tricks, and find the DE and DRNL to perform best on the training set.

Moreover, we find SEAL to perform very well at different rates. However, we do get a general trend of optimal performance for a learning rate of $1e-4$ and 2 hops. For the whole brain graph, we find Double-Radius Node Labeling (DRNL) to better capture the hierarchical structure \cite{zhang2018link}. Further, a DGCNN outperforms a GCN.
When training on a small region of interest we observe a gradual improvement for up to 50 epochs, for the whole brain graph we discover that SEAL almost converges the first epochs while inducing significant inductive bias, making it the superior model for our spatial and hierarchical graph. The training quickly plateaus. In all our experiments, the SEAL method performs best amongst all the baselines. All performance measures are given in Table \ref{tab:results_link_pred} in the main paper.

\subsection{Node classification}
\label{Suppl_Node_Class}

\begin{table}[h]
\centering
\caption{\label{tab:node_hyper_parameters} Details of the hyper-parameter search for node classification with the final parameters selected for the our baseline models.}
\tiny
\begin{tabular}{| l||l|l|l|}
\hline
Model                                         & Parameter Range & Selected Parameters & Model Select.\\
\hline 
Cluster-GCN                                     & lr $\in$ \{$2\cdot10^{-5}, 1\cdot10^{-4},
                                                      1\cdot10^{-3}, 3\cdot10^{-3}$\}           
                                                & lr = $3\cdot10^{-3}$
                                                & 1500 epochs   \\
                                                & number of layers $\in$ \{$3, 4, 5$\}                    
                                                & number of layers = $4$    
                                                &  \\
                                                & hidden channels $\in$ \{$128, 256, 512$\} 
                                                & hidden channels = $128$  
                                                &  \\
                                                & number of partitions $\in$ \{$3, 6, 9$\}  
                                                & number of partitions  = $9$ 
                                                &  \\
                                                & dropout $\in$ \{$0.0, 0.2, 0.5$\}  
                                                & dropout = 0.2 
                                                &  \\
\hline 
GNN                                             & lr $\in$ \{$2\cdot10^{-5}, 1\cdot10^{-4}, 1\cdot10^{-3},
                                                              3\cdot10^{-3}$\} 
                                                & lr = $3\cdot10^{-3} $ 
                                                & 1500 epochs   \\
                                                & number of layers $\in$ \{$3, 4, 5$\}                    
                                                & number of layers = $3$    
                                                &  \\
                                                & hidden channels $\in$ \{$32, 128, 256, 512, 1024$\} 
                                                & hidden channels = $256$ 
                                                &  \\
                                                & dropout $\in$ \{$0.1, 0.4, 0.5$\}  
                                                & dropout = $0.4 $
                                                &  \\
\hline 
GNN-SAGE                                        & lr $\in$ \{$2\cdot10^{-5}, 1\cdot10^{-4}, 1\cdot10^{-3},
                                                              3\cdot10^{-3}$\} 
                                                & lr = $3\cdot10^{-3}$  
                                                & $1500$ epochs   \\
                                                & number of layers $\in$ \{$2, 3, 4$\}                    
                                                & number of layers = $4$    
                                                &  \\
                                                & hidden channels $\in$ \{$32, 128, 256, 512, 1024$\}
                                                & hidden channels = $128 $
                                                &  \\
                                                & dropout $\in$ \{$0.0, 0.2, 0.4$\}  
                                                & dropout = $0.4$ 
                                                &  \\
\hline 
Graph-Saint                                     & lr $\in$ \{$1\cdot10^{-6}, 1\cdot10^{-5}, 1\cdot10^{-4}, 
                                                           1\cdot10^{-3}, 5\cdot10^{-3}, 1\cdot10^{-2}$\}
                                                & lr = $5\cdot10^{-4}  $
                                                & $1000$ epochs   \\
                                                & number of layers $\in$ \{$2, 3, 4$\}                    
                                                & number of layers = $4 $   
                                                &  \\
                                                & hidden channels $\in$ \{$64, 256, 512, 1024$\}                           
                                                & hidden channels = $64$ 
                                                &  \\
                                                & walk length $\in$ \{$3, 5, 7$\}   
                                                & walk length = $7$ 
                                                &  \\
                                                & dropout $\in$ \{$0.0, 0.35, 0.5$\}  
                                                & dropout = $0.35$
                                                &  \\
\hline 
SIGN                                            & lr $\in$ \{$1\cdot10^{-5}, 1\cdot10^{-4}, 1\cdot10^{-3},
                                                              5\cdot10^{-3}$\}
                                                & lr = $1\cdot10^{-3}$  
                                                & $1000$ epochs   \\
                                                & number of layers $\in$ \{$2, 3, 4$\}                    
                                                & number of layers = $3$    
                                                &  \\
                                                & hidden channels $\in$ \{$32, 64, 128, 256, 512, 1024$\}  
                                                & hidden channels = $128$ 
                                                &  \\
                                                & dropout $\in$ \{$0.0, 0.1, 0.3, 0.5$\}  
                                                & dropout = $0.1$
                                                &  \\

\hline 
MLP                                             & lr $\in$ \{$2\cdot10^{-5}, 1\cdot10^{-4}, 5\cdot10^{-4}, 
                                                           1\cdot10^{-3}, 3\cdot10^{-3}, 1\cdot10^{-2}$\}
                                                & lr = $1\cdot10^{-3} $ 
                                                & $1500$ epochs   \\
                                                & number of layers $\in$ \{$2, 3, 4$\}                    
                                                & number of layers = $3$   
                                                &  \\
                                                & hidden channels $\in$ \{$32, 128, 256, 512$\}
                                                & hidden channels = $256 $
                                                &  \\
                                                & dropout $\in$ \{$0.0, 0.3, 0.4$\}  
                                                & dropout = $0.0$
                                                &  \\
\hline 
SpecMLP-W + C\&S                                & lr $\in$ \{$1\cdot10^{-5}, 5\cdot10^{-4}, 1\cdot10^{-3}, 
                                                           3\cdot10^{-3}$\}
                                                & lr = $1\cdot10^{-3} $ 
                                                & $1500$ epochs   \\
                                                & number of layers $\in$ \{$3, 4, 5$\}                    
                                                & number of layers = $5$    
                                                &  \\
                                                & hidden channels $\in$ \{$128, 512, 1024$\}
                                                & hidden channels = $128 $
                                                &  \\
                                                & dropout $\in$ \{$0.0, 0.5$\}  
                                                & dropout = $0.5$
                                                &  \\
\hline
N2Vec                                           & lr $\in$ \{$1 \cdot10^{-5}, 1\cdot 10^{-3}, 
                                                              1 \cdot10^{-2}$\}
                                                & lr = $1 \cdot10^{-2}$\} 
                                                & $5$ epochs \\
                                                & walk length $\in$ \{$16, 40$\}           
                                                & walk length = $40$  
                                                &  \\
                                                & walks per node $\in$ \{$4, 10$\}      
                                                & walks per node  = $10$
                                                &  \\
                                                & embedding dim $\in$ \{$16,128$\}
                                                & embedding dim = $128$
                                                & \\
                                                & batch size $\in$ \{$16,128$\}
                                                & batch size = $128$
                                                & \\
                                                & epochs $\in$ \{$1, 5, 32$\}
                                                & epochs = $5$
                                                & \\
\hline
SpecMLP-W + C\&S + N2Vec                        & lr $\in$ \{$1\cdot10^{-5}, 5\cdot10^{-4}, 1\cdot10^{-3}, 
                                                           3\cdot10^{-3}$\}
                                                & lr = $1\cdot10^{-3} $ 
                                                & $1500$ epochs   \\
                                                & number of layers $\in$ \{$3, 4, 5$\}                    
                                                & number of layers = $5$    
                                                &  \\
                                                & hidden channels $\in$ \{$128, 512, 1024$\}
                                                & hidden channels = $128 $
                                                &  \\
                                                & dropout $\in$ \{$0.0, 0.5$\}  
                                                & dropout = $0.5$
                                                &  \\
\hline
\end{tabular}
\end{table}

Similar to the link prediction task, we employ a grid Search for hyperparameters. For each graph neural network, we explore the number of layers, hidden channels, learning rate and dropout ratio on the smaller sets validation and test set. For each model, if applicable, we select a set of hyper-parameters specific to the architecture and thus optimize. We selected our models according to the highest ROC AUC score on the validation set. 

The Cluster GCN algorithm provides a memory efficient alternative to other graph learning algorithms and is specifically designed to handle large-scale graphs efficiently. The number of partitions constitutes the essential hyper-parameter. We find it to consistently outperform other models, irrespective of its number of partitions.

Further we implement the base GCN by Kipf et.al. and extend it to the GNN GCN and GNN SAGE (SAGE introduces convolutional operator). For all we vary the number of hidden channels, number of layers and dropout ratio. In direct comparison, the GNN SAGE outperforms the GNN GCN by high margins (>10\%) for the ROC-AUC metric on the whole brain graph. Similar to the link prediction task, indicating that the base GCN is not suited for this task.

For the Node2Vec embeddings of the node classification task, we find similar trends for walk length, walks per node and embedding dimension as in the link prediction task. Please refer to Table \ref{tab:node_hyper_parameters} for the exact hyper-parameter selection.

For the MLP we select a comparative number of layers and obtain the best result for a high dropout ratio of 0.4.  We observe that the the model is unable to generalize to unseen data, indicating the absence of an appropriate inductive bias.

For the MLP-CS, we selected similar values in comparison to the MLP implementation. Incorporating the Correct\&Smooth methodology into the model increases the F1 score but does not improve the metrics which account for imbalanced datasets. 

The SIGN applies subsampling techniques. Similar to the graph neural network baselines, we explore the hidden channels and number of layers. We obtain the best for a high model complexity with 4 layers and 512 hidden channels. The model converges very fast with a high learning rate but performs only on par with the base GCN.

\subsection{Graph explainability}

We use the concept of the GNNExplainer \cite{ying2019gnnexplainer} as an initial graph explainability approach. GNNExplainer aims to identify a compact subgraph with a small subset of node features which play a crucial role in a given GNN for node classification. 

\begin{figure}[htp!]
    \centering
    \includegraphics[width=1.05\textwidth]{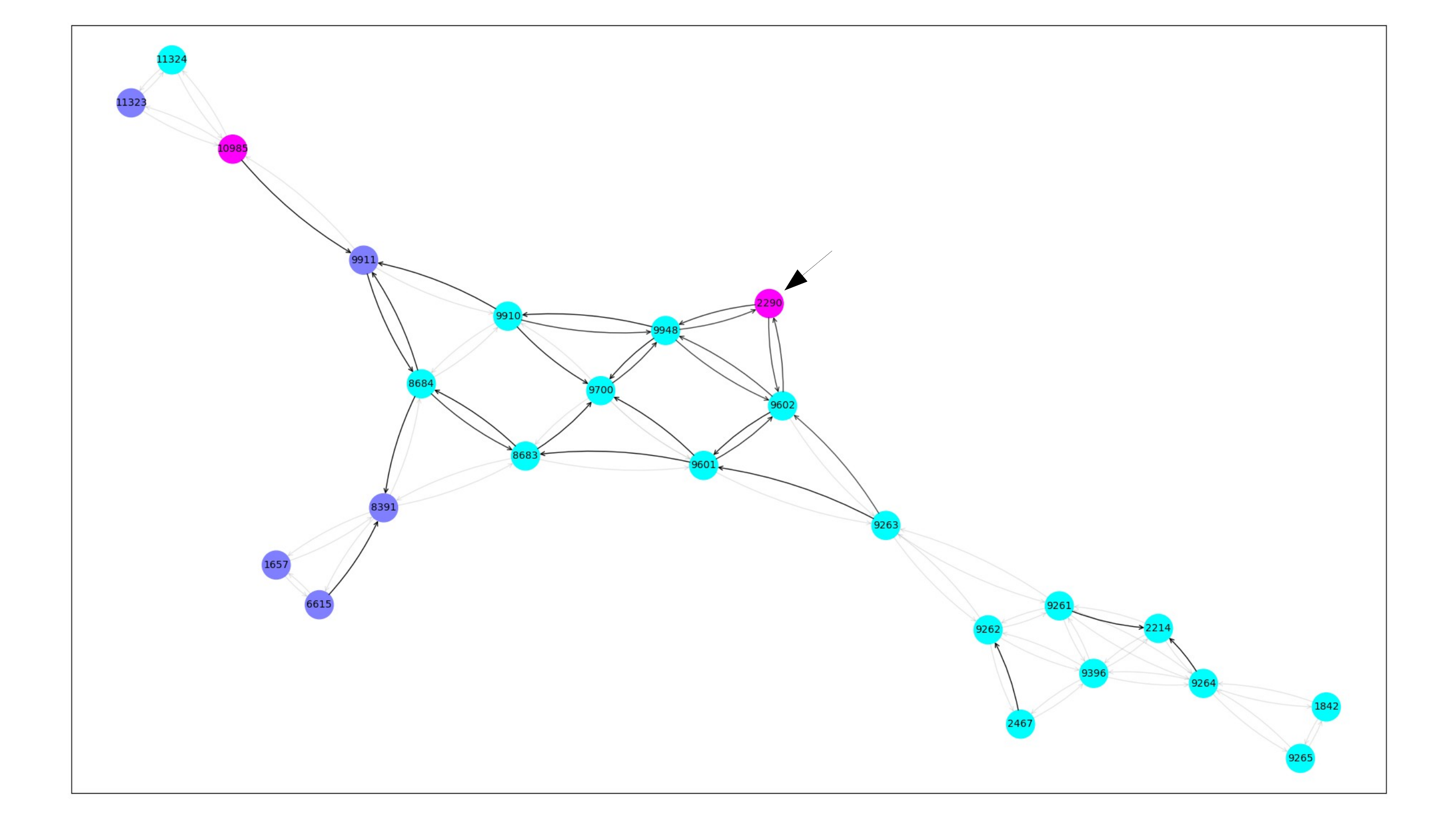}
    \caption{Graph explainability plot based on the GNNExplainer concept \cite{ying2019gnnexplainer}. Our considered node of interest is node $2290$. Pink nodes indicate nodes belonging artery/vein class, blue indicated arterioles/venules and green indicates capillaries. Thick edges represent a strong influence on the prediction. Please note that the considered graph is a spatial \textbf{Line Graph}.}
    \label{fig:graph_explain}
\end{figure}

We decided to run an initial GNNExpainer experiment on a small but representative subgraph of roughly $16000$ nodes and $18000$ edges.  We predict node 2290 with our trained SAGEConv based model for the node classification task on the Line Graph, see Figure \ref{fig:graph_explain}.  The visualization represents the nodes which our model considers important for the node prediction of $2290$. One can observe, that the model relies heavily on its immediate neighbourhood (denoted by thick edges). In our plot, we have visualized a 5 hop neighbourhood, for a SAGEConv GCN with 4 layers. Naturally, the model does not consider any node beyond its 4 hop neighbourhood in the computational graph.


\subsection{Computational resources}
\label{Comp_Resources}

All of our neural network trainings were performed on an Nvidia Quadro RTX 8000 GPU with 48GB memory.  

\clearpage

\section{Datasheet for datasets}
This description is an additional documentation intended to enhance reproducibility and follows the \textbf{Datasheets for Datasets}\footnote{\url{https://arxiv.org/abs/1803.09010}} working paper developed for the machine learning community. 

\begin{itemize}

\item \textbf{For what purpose was the dataset created?}
To foster research development in machine learning for graphs, in particular its application to neuroscience - specifically the brain vessel graph composition.

\item \textbf{Who created the dataset (e.g., which team, research group) and on behalf of which entity (e.g., company, institution, organization)?}
The dataset was created thorough a collaborative effort by neuro-scientists and computer-scientists at the Technical University of Munich and the Helmholtz Zentrum München (under the supervision of Ali Ertuerk, Bjoern Menze and Stephan Günnemann).
  
\item \textbf{Who funded the creation of the dataset?} 
The creation of the dataset was funded only indirectly via the salaries of the scientists at the Technical University of Munich and the other corresponding affiliations of the authors. 

\end{itemize}

\subsection{Composition}

\begin{itemize}

\item \textbf{What do the instances that comprise the dataset represent?} 
Our dataset represents graph representations of the whole brain vasculature. We are providing two alternative representations of the vascular graph. First, a representation where individual vessels are represented as edges in a Graph; and second, the corresponding Line graph were vessels are represented as nodes. One can interpret the graph of a single mouse brain as a single instance. Alternatively one can interpret each vessel (edge) and bifurcation (node) as a physical instance. 

\item \textbf{How many instances are there in total (of each type, if appropriate)?}
By the instance definition of whole brain graphs as instances we are providing 17 graphs (with the option to generate the line graph) from 3 imaging sources as instances. In the future we plan to extend the dataset as soon as other whole brain vessel segmentations are made publicly available (open source).  

By the definition of vessels and bifurcation points we have millions of instances for each. Please see Table \ref{tab:graph_numbers_nodes_edges} for detailed numbers.

\item \textbf{Does the dataset contain all possible instances or is it a sample of instances from a larger set?} We are providing all available instances. 

\item \textbf{What data does each instance consist of?} In either case, please provide a description. By definition 1); each instance represents a whole mouse brains' vascular graph saved in the widely used CSV format. By definition 2) each node represents a bifurcation point and each edge a vessel. 

\item \textbf{Is there a label or target associated with each instance?} Yes, in case of the edge and node instances, the information from the extracted graphs (features) can be used as the instance labels. E.g. in our node classification benchmark we use the vessel radius binned in three classes as an instance label. 
  
\item \textbf{Is any information missing from individual instances?}
No, all of the information has been provided.

\item \textbf{Are relationships between individual instances made explicit?} 
In our dataset the instances (brain graphs) are independent.
  
\item \textbf{Are there recommended data splits (e.g., training, development/validation, testing)?} 
For the benchmark we split one whole brain into a train, validation and test set of 80/10/10.
  
\item \textbf{Are there any errors, sources of noise, or redundancies in the dataset?} Our graph extraction is based on experimental imaging and segmentation techniques. Therefore, errors and uncertainty are inherent. We discuss these in detail in our Limitations section in the conclusion. 

\item \textbf{Is the dataset self-contained, or does it link to or otherwise rely on external resources (e.g., websites, tweets, other datasets)?} The provided dataset is self-contained. 

\item \textbf{Does the dataset contain data that might be considered confidential (e.g., data that is protected by legal privilege or by doctor-patient confidentiality, data that includes the content of individuals' non-public communications)?} No. 

\item \textbf{Does the dataset contain data that, if viewed directly, might be offensive, insulting, threatening, or might otherwise cause anxiety?} No. 

\item \textbf{Does the dataset relate to people?} No. 

\end{itemize}

\subsection{Collection process}

\begin{itemize}

\item \textbf{How was the data associated with each instance acquired?} The data was generated from a set of different publicly available datasets of whole murine brain images and segmentations. The specifics of the generation of each of these public segmentations are specified in the referenced literature and their licenses, see \ref{Individual_Licenses_Data}. 
    
\item \textbf{What mechanisms or procedures were used to collect the data?} We use the \textit{Voreen} framework \cite{drees2021scalable,meyer2009voreen} to generate graphs from segmentations. \textit{Voreen} is a software which runs on a CPU. 
    
\item \textbf{If the dataset is a sample from a larger set, what was the sampling strategy?} The dataset is complete. 

\item \textbf{Who was involved in the data collection process (e.g., students, crowdworkers, contractors) and how were they compensated (e.g., how much were crowdworkers paid)?}
Only researchers (co-authors) of the Technical University of Munich and the Helmholtz Zentrum München were involved in the data collection process.

\item \textbf{Over what timeframe was the data collected?} Does this timeframe match the creation timeframe of the data associated with the instances (e.g., recent crawl of old news articles)?  The generation of the dataset, including dedicated research to gather the base segmentations and to optimize the graph extraction procedure took roughly one year. 

\item \textbf{Were any ethical review processes conducted (e.g., by an institutional review board)?} Our work is purely based on public and open sourced data. However, ethical review processes were carried out for each of these open sourced base segmentation sets:

The three graphs from Ji et al. \cite{ji2021brain} are based on animal experiments,  they followed the Guide for the Care and Use of Laboratory Animals and have been approved by the Institutional Animal Care and Use Committee, for details see \url{https://doi.org/10.1016/j.neuron.2021.02.006}.

The animal experiments for the nine datasets from the VesSAP paper \cite{todorov2020machine} were carried out under approval of the ethical review board of the government of Upper Bavaria (Regierung von Oberbayern, Munich, Germany), and in accordance with European directive 2010/63/EU for animal research, for details see \url{https://doi.org/10.1038/s41592-020-0792-1}. 

\item \textbf{Does the dataset relate to people?} No.

\end{itemize}

\subsection{Preprocessing/cleaning/labeling}

\begin{itemize}

\item \textbf{Was any preprocessing/cleaning/labeling of the data done ?} Yes, this actually constitutes a core contribution of our work, therefore please refer to Section \ref{Graph_Ext} in the main paper and to Supplementary section \ref{Suppl_Graph_Doc}.

\item \textbf{Was the 'raw' data saved in addition to the preprocessed/cleaned/labeled data (e.g., to support unanticipated future uses)?} The raw data are the base segmentations. They are publicly available, the links are provided in Supplementary section \ref{Individual_Licenses_Data}.

\item \textbf{Is the software used to preprocess/clean/label the instances available?}
The \textit{Voreen} software used for the graph extraction is publicly available, see our github repo.
\end{itemize}

\subsection{Uses} 
\begin{itemize}
\item \textbf{Has the dataset been used for any tasks already?} In its current size and level of labeling detailization, the dataset was not used before (besides for the presented link prediction and node classification in this work). 

\item \textbf{Is there a repository that links to any or all papers or systems that use the dataset?} If so, please provide a link or other access point.
Yes, \url{https://github.com/jocpae/VesselGraph}.

\item \textbf{What (other) tasks could the dataset be used for?} In the main paper we discussed two standard tasks in machine learning on graphs; we think that our dataset can serve as a starting point for many interesting research directions in machine learning research and neurovascular research. 

\item \textbf{Is there anything about the composition of the dataset or the way it was collected and preprocessed/cleaned/labeled that might impact future uses?} No. 

\item \textbf{Are there tasks for which the dataset should not be used?} No. 
\end{itemize}

\subsection{Distribution}

\begin{itemize}

\item \textbf{Will the dataset be distributed to third parties outside of the entity (e.g., company, institution, organization) on behalf of which the dataset was created?} Our Dataset is open sourced under a CC Attribution-NonCommercial 4.0 International (CC BY-NC 4.0) License. Therefore all third parties can openly access it. 

\item \textbf{How will the dataset will be distributed (e.g., tarball  on website, API, GitHub)?} Yes, our DOI is \url{10.5281/zenodo.5301621}

\item \textbf{When will the dataset be distributed?} The dataset is available from the moment of submission. 

\item \textbf{Will the dataset be distributed under a copyright or other intellectual property (IP) license, and/or under applicable terms of use (ToU)?} Our Dataset is open sourced under a CC Attribution-NonCommercial 4.0 International (CC BY-NC 4.0) License.
\item \textbf{Have any third parties imposed IP-based or other restrictions on the data associated with the instances?} No. 
\item \textbf{Do any export controls or other regulatory restrictions apply to the dataset or to individual instances?} No.

\end{itemize}

\subsection{Maintenance}

\begin{itemize}

\item \textbf{Who is supporting/hosting/maintaining the dataset?}
The dataset is initially supported and maintained by the lead authors of this paper. The data is initially hosted on a university server and links are provided in the github repository \url{https://github.com/jocpae/VesselGraph}. In the long term we aim to incorporate our dataset into the open graph benchmark (OGB) initiative\footnote{\url{https://ogb.stanford.edu/}}. 

\item \textbf{How can the owner/curator/manager of the dataset be contacted (e.g., email address)?} Of course via e-mail:  \url{johannes.paetzold@tum.de} and via the github repository, see question above. 

\item \textbf{Is there an erratum?} At this stage no, but we are happy to track them in a dedicated file in our github repository. 

\item \textbf{Will the dataset be updated (e.g., to correct labeling errors, add new instances, delete instances)?} Yes, we release the dataset on open platforms on which we plan to continuously update our dataset. Particularly to add novel whole brain vessel graphs to the dataset.
    
\item \textbf{If the dataset relates to people, are there applicable limits on the retention of the data associated with the instances?} The dataset does not relate to people. 
    
\item \textbf{Will older versions of the dataset continue to be supported/hosted/maintained?} When novel versions of the dataset will be released we will continue to host and maintain the old versions of the dataset.
    
\item \textbf{If others want to extend/augment/build on/contribute to the dataset, is there a mechanism for them to do so?} We encourage other researches to exactly that. Depending on their contribution they can contribute to our github repository (in case of implementations) or reach out to us via e-mail in case they want to contribute graphs to the dataset. Our dataset and code are open sourced, see above. 

\end{itemize}

\end{document}